\documentclass[sigconf]{acmart} 
\usepackage{booktabs} 
\usepackage{balance} 

\usepackage{enumitem}
\usepackage{amsmath}
\usepackage{graphicx}
\usepackage{extarrows}
\usepackage{subfigure}
\usepackage{verbatim}
\usepackage{makecell}
\usepackage{diagbox}
\usepackage{bm}
\usepackage{graphics}
\usepackage[export]{adjustbox}
\usepackage{multirow}
\usepackage{float}
\usepackage{placeins}
\usepackage[linesnumbered, ruled]{algorithm2e}
\usepackage{epstopdf}
\usepackage{setspace}
\usepackage[super]{nth}
\usepackage{fancyhdr}
\usepackage{slashbox}
\usepackage{lipsum,multicol}
\usepackage{url}
\usepackage{xcolor}
\usepackage{tikz}
\usepackage[skip=3pt]{caption}
\usepackage{graphbox}

\usepackage[utf8]{inputenc}
\usepackage{array}
\usepackage{makecell}
\usepackage[compact]{titlesec} 

\def \eg {\emph{e.g.}, }
\def \ie {\emph{i.e.}, }

\allowdisplaybreaks[4]

\newcommand{\model}{{MOCHA}\xspace}
\allowdisplaybreaks[4]

\acmYear{2025}
\copyrightyear{2025}
\acmConference[SenSys '25]{The 23rd ACM Conference on Embedded Networked Sensor Systems}{May 6--9, 2025}{Irvine, CA, USA}
\acmBooktitle{The 23rd ACM Conference on Embedded Networked Sensor Systems (SenSys '25), May 6--9, 2025, Irvine, CA, USA}
\acmDOI{10.1145/3715014.3722044}
\acmISBN{979-8-4007-1479-5/25/05}

\AtBeginDocument{%
  \providecommand\BibTeX{{%
    \normalfont B\kern-0.5em{\scshape i\kern-0.25em b}\kern-0.8em\TeX}}}

\begin{document}

\title{Responsive DNN Adaptation for Video Analytics against Environment Shift via Hierarchical Mobile-Cloud Collaborations
}

\author{Maozhe Zhao}
\orcid{0009-0004-5492-0897}
\affiliation{
 \institution{Shanghai Jiao Tong University}
 \city{Shanghai}
\country{China}
 }
\email{larval_roc@sjtu.edu.cn}

\author{Shengzhong Liu}
\orcid{0000-0002-7643-7239}
\authornote{Shengzhong Liu is the corresponding author.}
\affiliation{
 \institution{Shanghai Jiao Tong University}
 \city{Shanghai}
\country{China}
 }
\email{shengzhong@sjtu.edu.cn}

\author{Fan Wu}
\orcid{0000-0003-0965-9058}
\affiliation{
 \institution{Shanghai Jiao Tong University}
 \city{Shanghai}
\country{China}
 }
\email{fwu@sjtu.edu.cn}

\author{Guihai Chen}
\orcid{0000-0002-6934-1685}
\affiliation{
 \institution{Shanghai Jiao Tong University}
 \city{Shanghai}
\country{China}
 }
\email{gchen@sjtu.edu.cn}

\renewcommand{\shortauthors}{Maozhe Zhao et al.}

\begin{abstract}
Mobile video analysis systems often encounter various deploying environments, where environment shifts present greater demands for responsiveness in adaptations of deployed ``expert DNN models''.
Existing model adaptation frameworks primarily operate in a cloud-centric way, exhibiting degraded performance during adaptation and delayed reactions to environment shifts. 
Instead, this paper proposes \model, a novel framework optimizing the responsiveness of continuous model adaptation through hierarchical collaborations between mobile and cloud resources.
Specifically, \model 
(1) reduces adaptation response delays by performing on-device model reuse and fast fine-tuning before requesting cloud model retrieval and end-to-end retraining;
(2) accelerates history expert model retrieval by organizing them into a structured taxonomy utilizing domain semantics analyzed by a cloud foundation model as indices;
(3) enables efficient local model reuse by maintaining onboard expert model caches for frequent scenes, which proactively prefetch model weights from the cloud model database.
Extensive evaluations with real-world videos on three DNN tasks show \model improves the model accuracy during adaptation by up to 6.8\% while saving the response delay and retraining time by up to $35.5\times$ and $3.0\times$ respectively.
\end{abstract}



\maketitle

\section{Introduction}\label{sec:intro}
Video analytics~\cite{10.1145/3570361.3615749,295529,DBLP:conf/mobicom/ShiLJHHZYBX0YX24,9912287} with deep neural networks (DNN) have been extensively applied in Internet of Things (IoT) applications, especially autonomous driving~\cite{10.1145/3495243.3560539,10.1145/3570361.3613280,10.1145/3560905.3568519} and low-cost intelligent robots~\cite{DBLP:conf/icarm/WuDXW0024,10.1145/3666025.3699325,10.1145/3699779}. 
Mobile video analytics perform DNN-based inference on mobile devices to deliver real-time results~\cite{10.1145/3560905.3568546,10.1145/3625687.3625789,xu2024flex}.
Due to the resource constraints on mobile devices, dedicated \textit{expert models}, with compressed size and distilled knowledge from the large-scale \textit{teacher model}, are deployed to balance the prediction fidelity and processing throughput~\cite{gui2019model,295531,10.1145/3560905.3568520}.
This makes the analytics system responsive and accurate to data that matches the expert model's training distribution.

However, numerous tasks are not confined to a single scene. They are subject to various unpredictable domain shifts~\cite{suprem13odin,luo2019taking}, which alter the data distribution of collected video streams. In this paper, we focus on environment shifts, mainly referring to shift factors related to the natural environment, which is a type of domain shift and can easily be recognized in real-world applications.
For example, light conditions can change dramatically when vehicles pass through consecutive tunnels, and backgrounds can shift significantly when re-entering elevators or buildings.
In these environment shifts, while teacher models exhibit robustness, they cannot be directly applied to lightweight mobile devices. Expert models, on the other hand, can only memorize limited video scenes (\ie domain), resulting in poor generalization in environments deviating from their training domain~\cite{10.1145/3625687.3625800}.
Furthermore, the safety-critical nature of these tasks requires analytics systems to maintain high accuracy throughout their operations. 
A slight performance degradation at any time can impact the quality of service (QoS). Unlike regular working conditions, environment shifts put additional pressure on the systems' accuracy and must be managed precisely.

To sustain stable DNN accuracy against environment shifts, continuous expert model adaptations are conducted in previous frameworks~\cite{DBLP:conf/sensys/GaoTCC0H24,DBLP:conf/sensys/JiZ24,DBLP:conf/sensys/KaraKC00CWLA24,DBLP:conf/sensys/LiLL0024}, where two strategies are commonly exploited :
(1) \textit{Model retraining}~\cite{parisi2019continual,beaulieu2020learning,fang2024adashadow} updates the expert model parameters with freshly collected samples;
(2) \textit{Model reuse}~\cite{khani2023recl} compares and selects one best candidate from a history model zoo to replace the obsolete one.
However, due to resource-intensive computation or storage needs, existing solutions typically host model adaptation services on the cloud~\cite{suprem13odin,khani2023recl,bhardwaj2022ekya}. This produces delayed responses (\ie process of completing the adaptation and returning to high accuracy) to environment shifts:
First, cloud-based model adaptation requires frequent data and model weight exchange (communication overhead time) between mobile and cloud~\cite{10.1145/3560905.3568538,10.1145/3485730.3485929,10.1145/3503161.3548033}, leading to considerable communication delays and bandwidth consumption;
Second, centralized adaptation paradigms make the cloud a single point of congestion, leading to poor scalability to the number of mobile devices with long cloud queuing delays.
In tasks involving multiple environments, environment shifts are inevitable and such shifts make QoS unacceptable with delayed responses. If DNN-based analytics systems can recover more quickly with stronger responsiveness by reducing the aforementioned overheads, they will achieve higher overall accuracy during environment shifts, agnostic to the model optimization algorithms used.

This paper optimizes the \textit{response efficiency} of continuous model adaptation in mobile video analytics against environment shifts by efficiently exploiting mobile resources through a novel mobile-cloud collaborative paradigm, thereby improving performance in autonomous driving.
The key intuition is to leverage lightweight mobile adaptation actions to reduce the frequency of requesting cloud adaptation services and avoid unnecessary mobile-cloud communications. 
Two technical challenges are raised:
First, considering the cloud network delays and mobile resource constraints, how to design a mobile-cloud collaboration algorithm taking advantage of both the low latency on the mobile and rich resources on the cloud is complex. We need to rethink a brand-new arrangement of inference and various adaptations across mobile devices and cloud servers, one that differs from all previous systems. The integration of these design components is, in itself, a tough challenge.
Second, forcibly moving intact model reuse and retraining functions from cloud servers to mobile devices induce excessive computation and storage load to onboard resources. We need to identify suitable meta-level information to reduce the algorithm's computational demand while ensuring accuracy and responsiveness.

We introduce \model, a mobile-cloud collaborative model adaptation framework for mobile video analytics, which achieves the objective without introducing a new algorithm. 
It distributes hierarchical adaptation functions between mobile and cloud resources.
We serve most model reuse and retraining requests locally on mobile through retrieval from a small expert model cache and updating the expert model with single-layer LoRA (Low-Rank Adaptation) fine-tuning while offloading intensive requests to the cloud server which hosts a full-fledged expert model database and runs end-to-end expert model retraining.
The cloud only works as a backup to the mobile when lightweight onboard services do not suffice to achieve the adaptation objective. 

The design of \model includes three key components.
First, \model offloads lightweight model adaptations to the mobile devices for better system responsiveness, which composes a two-tier adaptation hierarchy. It applies on-device model reuse and fast LoRA single-layer fine-tuning as immediate actions against environment shifts, before data transmission to the cloud for cloud model retrieval and end-to-end retraining.
Second, it constructs a structured model database organized by domain semantics, which can be analyzed by a foundation model (FM) on the cloud, to retrieve the fittest model candidate among history models in fast adaptations.
Third, \model maintains onboard model caches to store model weights for popular scenes, enabling seamless model switching with minimal onboard transmission delays. The mobile cache proactively interacts with the cloud model database for prefetching to ensure uninterrupted inference.

We evaluate MOCHA with two large-scale real-world video datasets on three analytics tasks: object detection with YOLO~\cite{jocher2021ultralytics}, image classification with ResNet~\cite{he2016deep}, and semantic segmentation with Deeplabv3+~\cite{chen2018encoder}. We extensively compare it against state-of-the-art (SOTA) model adaptation systems.
The results show that \model supports more devices than SOTA methods in lower response delays with the same cloud resources. As the mobile device number scales, \model improves the accuracy during adaptation by up to 6.8\%, and saves the response delay and retraining time by $35.5\times$ and $3.0\times$. Ablation results validate that \model's superior performance mainly comes from the onboard adaptations of fine-tuning and reuse with mobile model cache.

In summary, the main contributions of this paper are:
\begin{itemize}[topsep=0pt,leftmargin=0.35cm]
    \item We propose a mobile-cloud collaborative continuous adaptation framework \model for effective mobile video analytics against runtime scene changes.
    \item We design an adaptation hierarchy of onboard model reuse, fine-tuning, and cloud retraining for better 
    responsiveness.
    \item We introduce an automated semantic model taxonomy for real-time domain metadata-based model retrieval used within adaptation strategies.
    \item We implement \model and perform extensive evaluations with a real-world video dataset, where \model saves the adaptation latency by up to $35.5\times$.
\end{itemize}

\section{Background and Motivations} \label{sec:motivation} 

\begin{figure}[t!]
\includegraphics[width=0.95\linewidth]{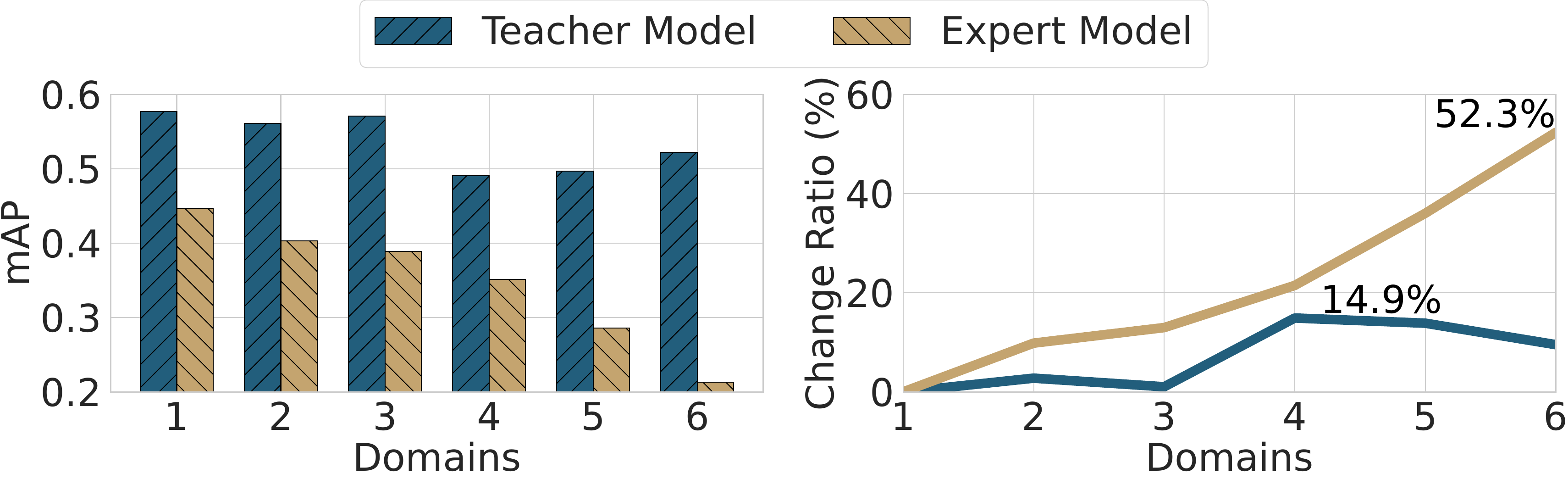}
\caption{Teacher Model vs. Expert Model. The change ratio denotes the ratio of accuracy drop in the current domain over accuracy in the optimal domain, reflecting model generalizability in different environments.}
\label{fig:golden_vs_expert}
\end{figure} 

Here, we introduce the background of continuous adaptations against frequent environment shifts in mobile video analytics (§\ref{background:continuous_video_analytics}), analyze the advantages of the hierarchical model adaption (§\ref{background:mobile_cloud_collab}), and give intuitions of semantics-indexed model retrieval (§\ref{background:semantic_index}).

\subsection{Continuous Adaptation against Frequent Environment Shift} \label{background:continuous_video_analytics}

\begin{figure}[t!]
\includegraphics[width=0.95\linewidth]{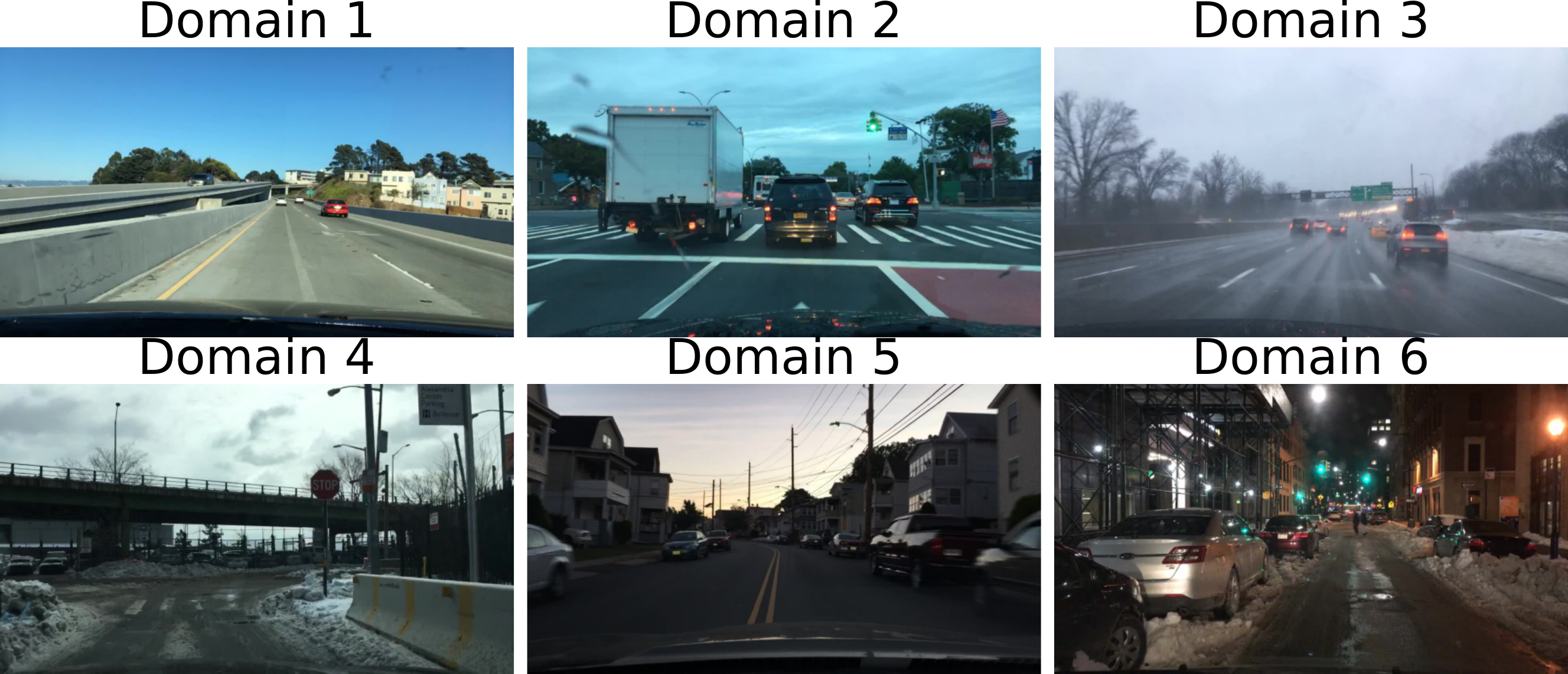}
\caption{Example frames of the domains in Figure~\ref{fig:golden_vs_expert}.}
\label{fig:domain_show}
\end{figure}

Environment shifts in video analytics are common, particularly in mobile systems. In the datasets we collected, we find that numerous environment shifts occur, lowering the system's performance. Compared to the teacher model, expert models generalize poorly against environment shifts. As Figure~\ref{fig:golden_vs_expert} and Figure~\ref{fig:domain_show} show, the accuracy degradation of the teacher model remains within 15\% across different domains, while the expert model experiences up to a 52\% accuracy drop. Therefore, during frequent environment shifts, continuous adaptations are needed to stabilize DNN accuracy.

\begin{figure}[t!]
\includegraphics[width=0.95\linewidth]{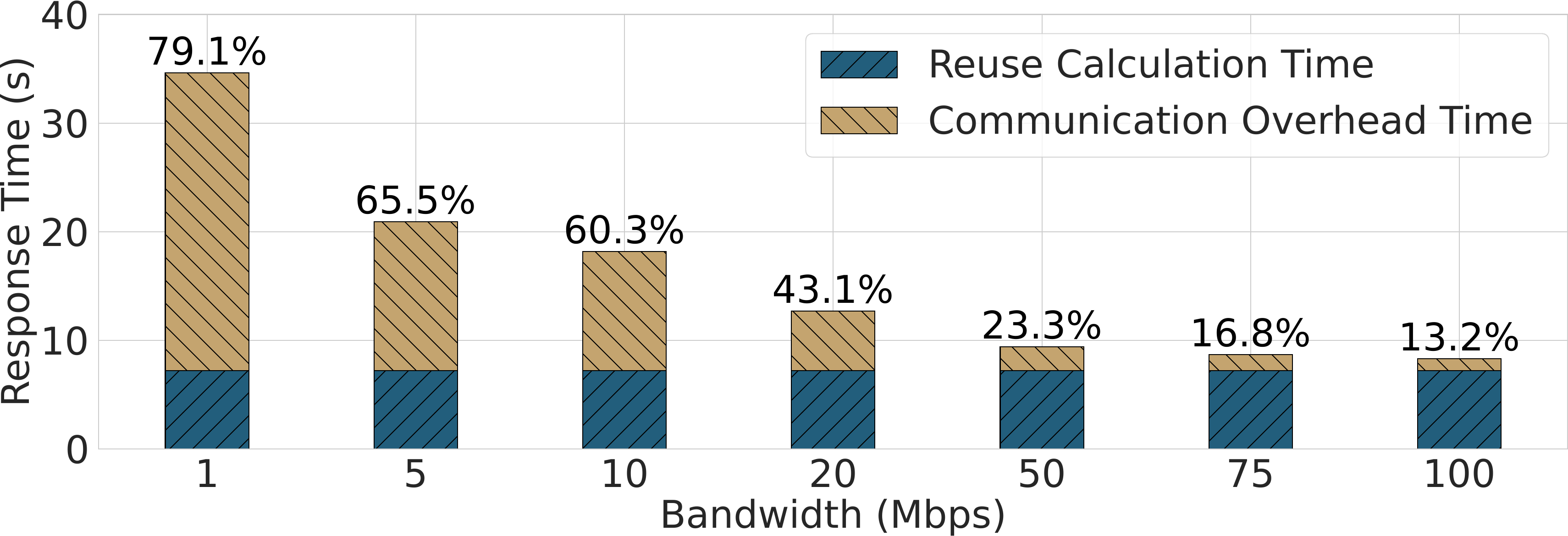}
\caption{Communication overhead ratio comparison. The system first calculates the model to reuse and then transmits model weights to the mobile device. }
\label{fig:communication_overhead}
\end{figure}

However, the two adaptation strategies (\ie reuse and retraining) face distinct challenges. 
Reuse requires a trade-off between response speed and accuracy by making different choices of model selection within it while retraining needs more compute resources and cannot provide a responsive solution at once~\cite{DBLP:conf/sensys/BarjamiMM24}. 
To guarantee DNN accuracy, most existing solutions~\cite{suprem13odin,khani2023recl,bhardwaj2022ekya,khani2021real} completely offload these adaptations to the remote cloud server, leading to suboptimal response efficiency during environment shifts, due to mobile-cloud communication delays and cloud queuing delays~\cite{10.1145/3485730.3492885}. As shown in Figure~\ref{fig:communication_overhead}, the communication overhead can account for over 60\% of the total response time when the bandwidth is 10 Mbps, highlighting the inadequacy of cloud-based strategies to handle frequent environment shifts.
To improve the responsiveness, our intuition is to deploy partial actions to mobile devices and design a hierarchical adaptation system~\cite{DBLP:conf/mobicom/LiLLLCZWWLL24,9381477}.

\begin{table}[t!]
\caption{Resource metrics of different tasks. Bandwidth is 10 Mbps and model weight size is 14 MB.}
\resizebox{0.95\linewidth}{!}{
\begin{tabular}{c|ccc}
\toprule
Task         & Speed (FPS) & Total Time (s) & Memory (GB) \\ \midrule
Inference    & 8.3           & /       & 2.2        \\ \midrule
Onboard Fine-tune    & 9.7           & 120     & 2.6        \\
Cloud Retrain      & 303           & 160     & 14.4       \\ \midrule
Onboard Reuse & /             & 0.47     & /          \\
Cloud Reuse  & /             & 22.8    & /          \\ \bottomrule
\end{tabular}
}
\label{tab:task-cost}
\end{table}


\subsection{Hierarchical Model Adaptation}\label{background:mobile_cloud_collab}

The ``hierarchy'' can be interpreted from two orthogonal perspectives: 
First, model adaptation tasks can be defined hierarchically: model reuse, fine-tuning a few layers, and end-to-end retraining, with increasing adaptability but decreasing efficiency.
Second, the model adaptation could happen collaboratively between mobile devices and the cloud for balanced response efficiency and adaptation effectiveness.

\subsubsection{\textbf{Mobile Fine-tuning vs. Cloud Retraining}.} 
The benefits of model reuse in reducing model retraining have been highlighted in RECL~\cite{khani2023recl}, while we further address the importance of lightweight model fine-tuning that only selectively updates a few layers of a DNN. 
It is more computationally efficient than end-to-end retraining and attains better adaptability than reuse when no suitable history expert model can be directly picked.
As can be seen from Table~\ref{tab:task-cost}, although onboard fine-tuning takes longer per iteration, its total update time is shorter due to the reduced training iterations with fewer data and the absence of communication and queuing delays encountered in cloud retraining tasks.
Based on the extent of environment shift, the combination of reuse, fine-tuning, and retraining constitutes a more fine-grained action space for model adaptation.

\subsubsection{\textbf{Mobile Reuse vs. Cloud Reuse}.}
The limited mobile storage and computation capacity have not been sufficiently explored in model adaptations, where model reuse can demonstrate its value.
Intuitively, we make an analogy between mobile-vs-cloud resources and the hierarchical storage of computer systems. 
We have limited but fast-accessing resources (``cache'') on mobile devices, and abundant but distant resources (``main memory'') on the cloud server. 
From the storage aspect, saving a few frequent expert models on the mobile can save remote access delays.
From the computation aspect, distributing lightweight fine-tuning on mobile devices while centralizing heavyweight retraining on the cloud can balance the response efficiency and processing throughput in expert model updates.

\begin{figure}[t!]
\includegraphics[width=0.9\linewidth]{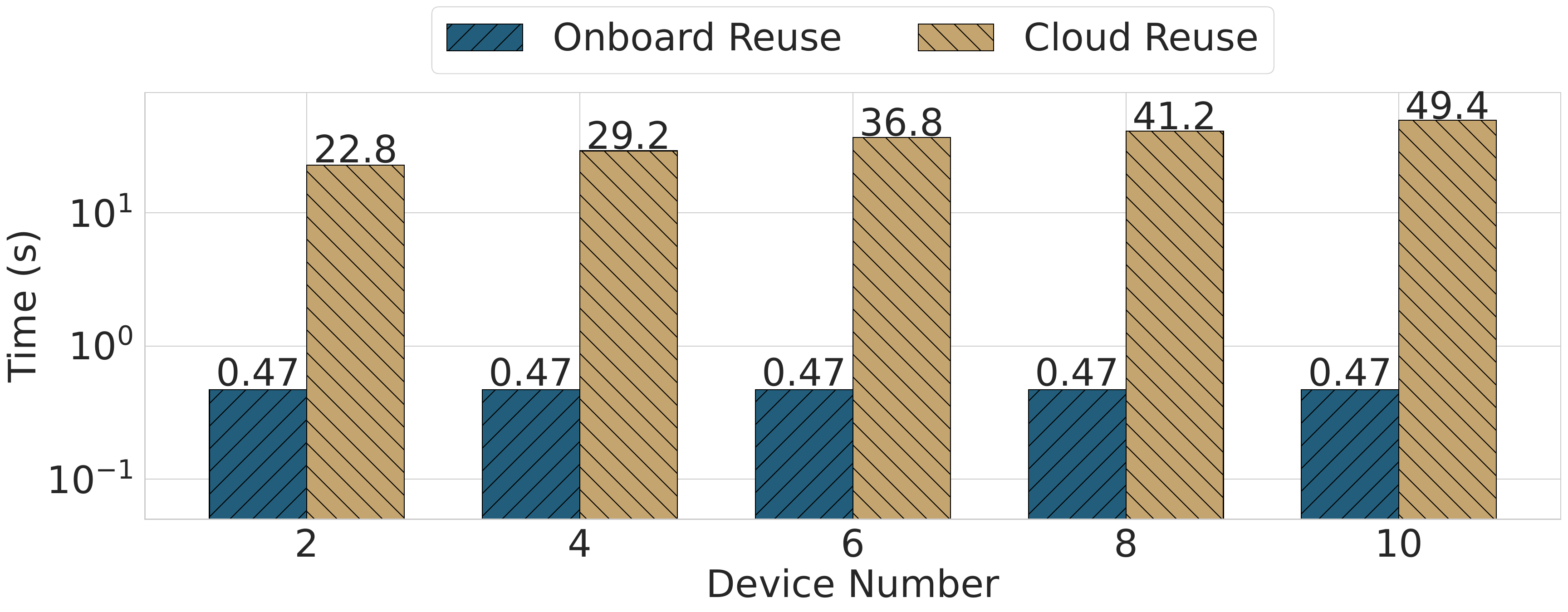}
\caption{Onboard Reuse v.s. Cloud Reuse.}
\label{fig:reuse-time}
\end{figure}

For example, in object detection tasks, a YOLOv5-s model (17M FLOPs) weights file is about 14MB, thus we can store multiple expert models on the mobile device. 
As shown in Table~\ref{tab:task-cost}, if the expert model is cached onboard, we can directly load it in 0.47s, without waiting for the cloud dispatch, thus improving the overall response speed~\cite{xue2024powerinfer}.
We further analyze the scalability of mobile device numbers between local reuse and cloud reuse. As Figure~\ref{fig:reuse-time} shows, onboard retrieval exhibits magnitudes shorter response delays than cloud model retrieval, which gap is further pronounced when more mobile devices need to be served by the cloud.

\begin{table*}[t!]
\caption{Examples of foundation model (FM)-based domain semantic discrimination.}
\label{tab:FM}
\resizebox{0.9\textwidth}{!}{
\begin{tabular}{c|lll}
\toprule
\textbf{Prompt}    & \multicolumn{3}{l}{\begin{tabular}[c]{@{}l@{}}Identify the \textit{location}, the \textit{weather}, and the \textit{time} in the image. \\ Respond in three words in the order of location, weather, and time, separated by commas, all lowercase.\end{tabular}} \\ \midrule
\textbf{Images}    & \multicolumn{1}{c|}{\includegraphics[align=c, width=0.25\textwidth]{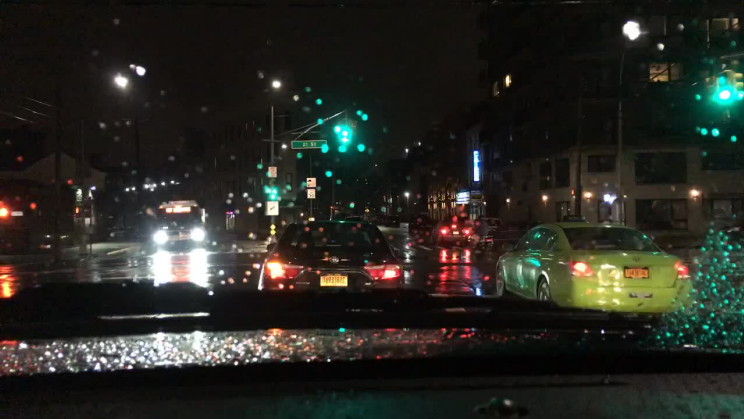}}     & \multicolumn{1}{c|}{\includegraphics[align=c, width=0.25\textwidth]{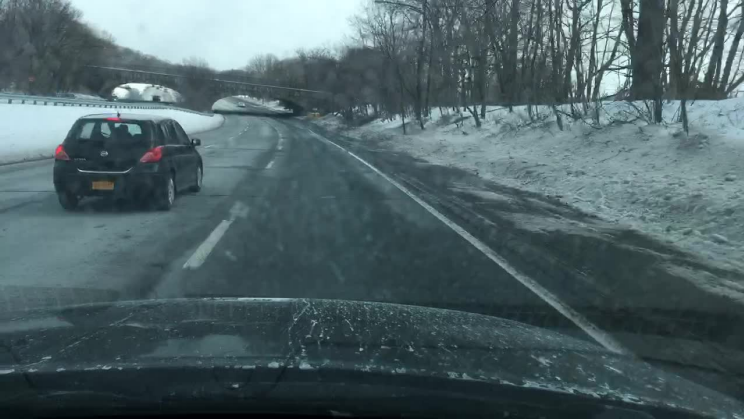}}     & \multicolumn{1}{c}{\includegraphics[align=c, width=0.25\textwidth]{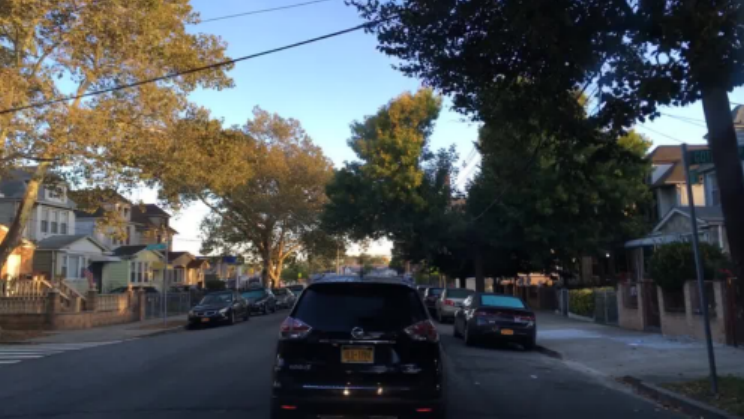}}    \\ \midrule
\textbf{Responses} & \multicolumn{1}{c|}{street, rainy, night}                                                                 & \multicolumn{1}{c|}{highway, snowy, daytime}                                                                       & \multicolumn{1}{c}{residential, clear, daytime}                                                                \\ \bottomrule
\end{tabular}
}

\end{table*}



\begin{figure*}[t!]
\centering
\begin{minipage}{.45\linewidth}
  \centering
    \includegraphics[width=0.95\linewidth]{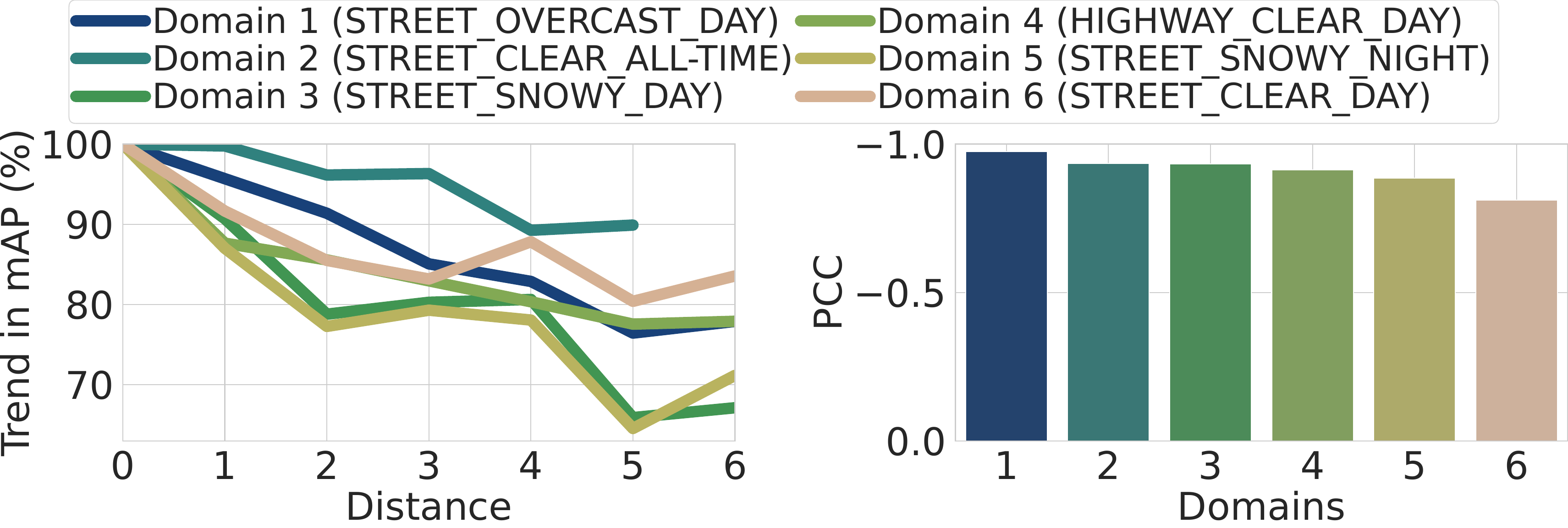}
    \caption{Reuse on similar domains. Pearson correlation coefficient (PCC) is calculated from the distance and mAP for each validation domain.} 
    \label{fig:similarity_reuse}
\end{minipage}
\begin{minipage}{0.03\linewidth}
\
\end{minipage}
\begin{minipage}{.45\linewidth}
  \centering
    \includegraphics[width=0.95\linewidth]{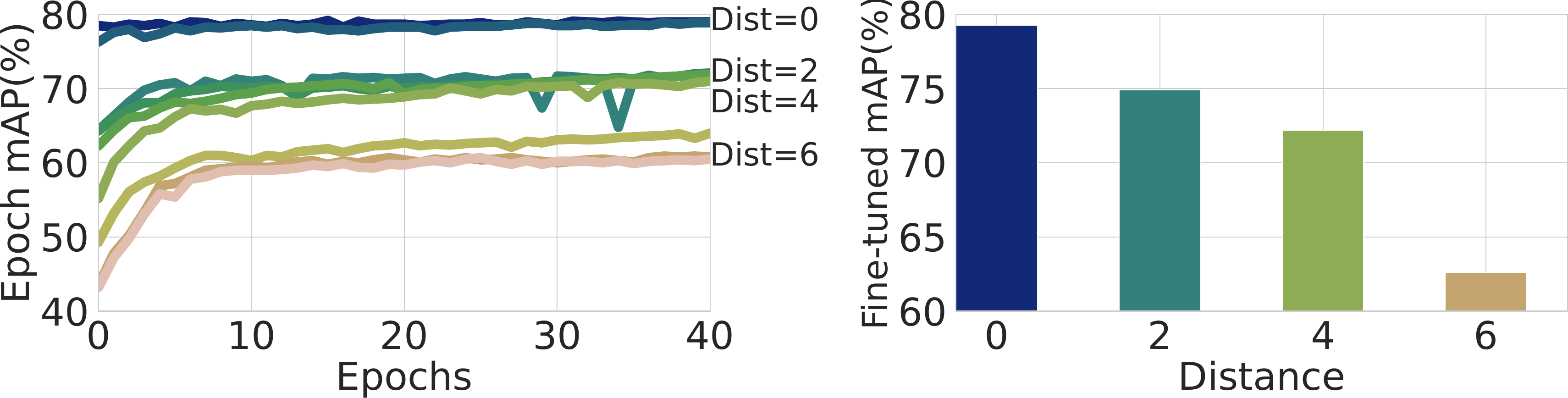}
    \caption{Fine-tuning on similar domains. Models from domains at varying distances are used as starting points for fine-tuning on the same dataset.}
    \label{fig:similarity_finetune}
\end{minipage}
\vspace{-0.4cm}
\end{figure*}

\subsection{Semantic-Indexed Model 
Retrieval}\label{background:semantic_index}
One critical step to hierarchical adaptations is the efficient retrieval of the best candidate for the new domain from history expert models. 
Besides iterating over all models with a benchmark dataset, recent studies have employed unsupervised clustering~\cite{suprem13odin} or maintained a gate network~\cite{khani2023recl} for expert model selection. However, these learnable methods require frequent updates when the model set changes.
We argue that they all overlook recognizable but critical semantic information of the deployed environments~\cite{DBLP:conf/sensys/JainMB24}, which can be used to organize history expert models into a structured database, instead of a disorganized model zoo, to improve the model retrieval efficiency.

\subsubsection{\textbf{Domain Semantics Recognition.}}
Domain semantics refer to the environmental attributes that describe and differentiate domains in environment shifts. \footnote{The selection of attributes' dimensions is orthogonal to this paper, which is hard to analyze quantitatively. We assume users provide parts of these parameters and the system can complete the others and construct the taxonomy offline using a few-shot validation dataset.} This includes factors like time, weather or location.
These domain semantics can be easily recognized by general foundation models (FM) in a zero-shot way~\cite{10.1145/3666025.3699331,10590439,DBLP:conf/sensys/SrivastavaKP0J24}, which has been pre-trained on massive data and can be used for diverse downstream tasks. Existing FMs, like LLaVA ~\cite{liu2023llava}, have been pre-trained to comprehend multi-modal input by connecting image input with a text prompt~\cite{DBLP:conf/sensys/LuD0FZTW0X024}. In Table \ref{tab:FM}, we use LLaVA to recognize the time, weather, and location of an image with a dedicated prompt. It achieves promising accuracy in discriminating domains. The computation for a $640\times640$ image takes 500ms on an NVIDIA RTX4090 GPU. 
We request human operators to provide a subset of domain semantic dimensions with value options as a one-time effort to facilitate expert model management, while other uninterpretable factors are implicitly handled in model continual learning.

\subsubsection{\textbf{Domain Semantics Utilization.}} \label{subsubsec:semantics}
The semantic dimensions can build a hierarchical taxonomy and a distance space on different domains, which can be directly used to retrieve the fittest model candidate in fast adaptations.
We hypothesize that the semantic-level distance can work as an indicator of cross-domain model reuse and retraining effectiveness. 
We conduct a few experiments to analyze the cross-domain gaps for empirical validation.
Intuitively, two domains are similar if they share most attributes but differ in a few. If all attributes match, the domains are identical.

\textbf{Direct Reuse:} We use data from BDD100K~\cite{bdd100k} and a set of YOLOv5-s pre-trained models to perform object detection on more than 10 validation domains divided by three attributes ``location'', ``weather'', ``time\_of\_day'', and record the mean Average Precision (mAP) as detection accuracy. 
We use a distance (introduced in §\ref{subsec:taxonomy} of \model) as domain similarities. We list 6 domains of model performances in Figure~\ref{fig:similarity_reuse}. 
More similar domains tend to have higher cross-domain accuracy, and the highest accuracy is achieved in the training domain (\ie distance=0).
The semantic similarity exhibits a strong negative correlation with model performance, with the overall average Pearson correlation coefficient (PCC) exceeding -0.82.
With more shared attributes, domain distributions become closer or even overlapped. In the overlapped scenes, models from both domains produce similar outputs and close accuracy.

\textbf{Lightweight Fine-tuning:} We also test whether models from similar domains are more suitable for fast fine-tuning. We use the same data and models to conduct a fine-tuning task with fixed epochs while freezing all but a single layer of the model. 
In Figure~\ref{fig:similarity_finetune}, we fine-tune models from all domains on the ``STREET\_CLEAR\_DAY'' domain and test mAP on its validation data. The results show that more similar domains can also obtain faster convergence and better accuracy improvement after fine-tuning.
Thus, semantic similarity can also be used as an indicator for cross-domain fine-tuning.

\section{\model Framework} \label{sec:framework}

\subsection{Overall Architecture} \label{subsec:overall}
\begin{figure}[t!]
    \includegraphics[width=0.9\linewidth]{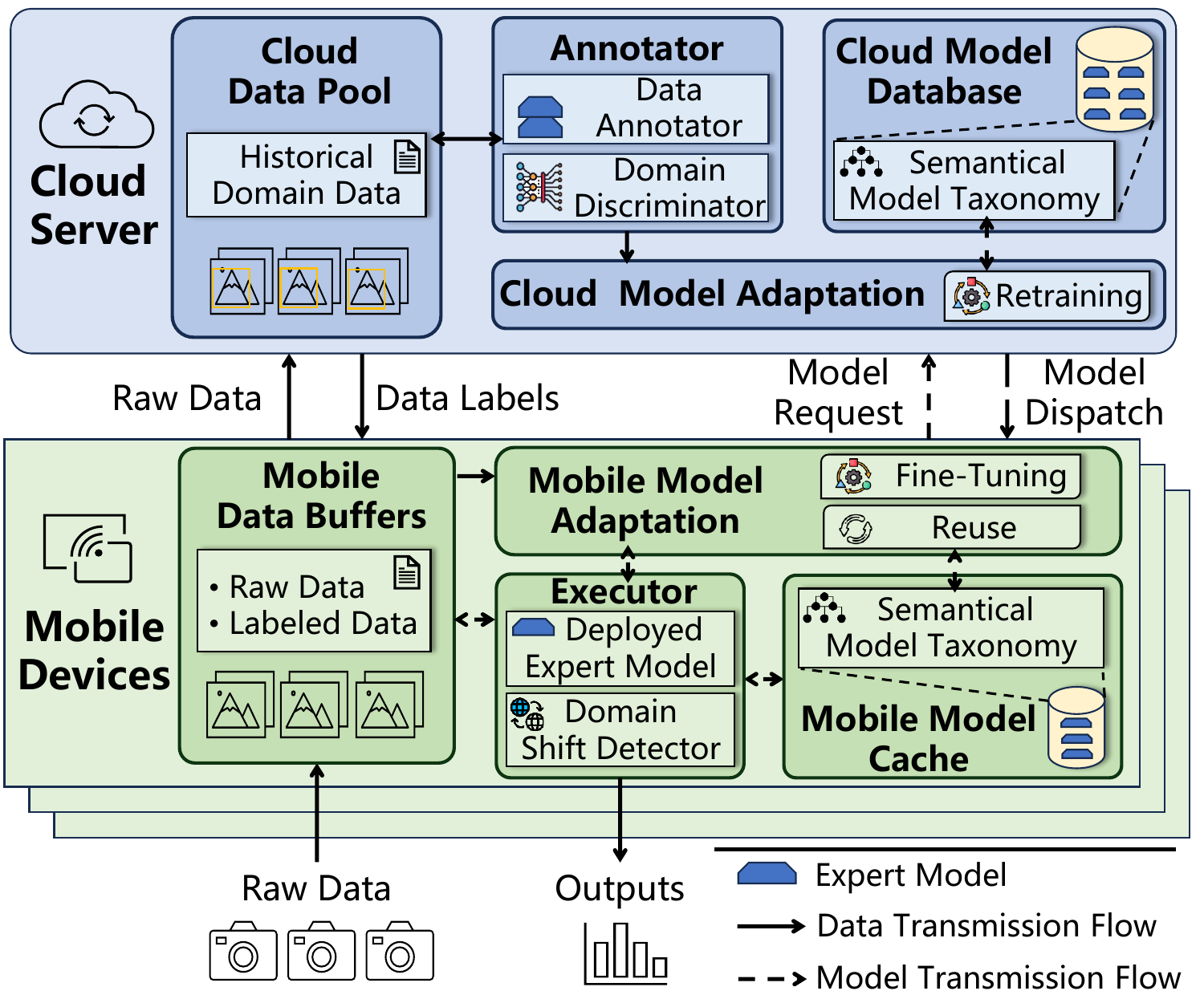}
    \caption{Architecture of \model. }
    \label{fig:architecture}
\end{figure}

As Figure~\ref{fig:architecture} shows, \model is a mobile-cloud collaborative model adaptation framework for continuous mobile video analytics with resilient prediction fidelity.
Upon an environment shift, the system should calibrate the mobile-deployed model to recover high accuracy within a short response latency.
We consider a cloud server with powerful compute and storage resources, and a set of distributed mobile devices with constrained compute and communication resources (\eg NVIDIA Jetson Nano, TX2).
\model offloads low-latency adaptation to mobile devices through lightweight computations (\ie local reuse and fast fine-tuning), with background support from the cloud server through heavyweight computations (\ie end-to-end retraining).

\textbf{Optimization Objective:}
The core tasks of continuous modal adaptation include ``\textit{detecting when the environment shift happens}'' and ``\textit{adapting the mobile-deployed expert model}'', optimizing the responsiveness (\ie finish adaptation in a short time) and scalability (\ie support more mobile devices) of adaptations, under the constraints of mobile resources and mobile-cloud network delays. 
\model performs environment shift detection on mobile devices, implements model-cache-based reuse and LoRA single-layer fine-tuning as immediate onboard reactions, and leaves accurate model retrieval from a full-fledged model database and end-to-end retraining on the cloud as backup strategies. 
\model's response efficiency benefits from its mobile-cloud collaborative nature on reuse and update strategies, as defined below.
\begin{itemize}[topsep=0pt,leftmargin=0.35cm]
    \item \textbf{Model Reuse:} If the exact or a similar domain has appeared in the past, we directly retrieve and use the previously trained model from the mobile model cache or the cloud model database as a response.
    \item \textbf{Model Update:} If no previous model can fit in the current domain, we update part or all of a model's parameters through lightweight onboard fine-tuning or end-to-end retraining on the cloud server to respond with delays.
\end{itemize}
\model's performance highly relies on effective and efficient history model retrieval from past domains.
\model leverages meta-level domain semantics identified by a multi-modal foundation model, as indices to construct a domain taxonomy for real-time expert model retrieval.
Below, we list the main mobile and cloud components in \model.

\subsubsection{\textbf{Mobile Components.}} 
Besides regular model inference, the mobile device periodically detects the occurrence of environment shifts and performs fast onboard model reuse and fine-tuning, through the following four modules. 
\begin{itemize}[topsep=0pt,leftmargin=0.35cm]
    \item \textbf{Executor}: It performs inference on the real-time video stream. Besides, a lightweight environment shift detector is integrated to detect potential distributional drifts. 
    \item \textbf{Mobile Model Cache}: It stores a few expert models that could be reused shortly. At the meta-level, it keeps a copy of the semantic model taxonomy (synchronized with the cloud server) to assist in model selection.
    \item \textbf{Mobile Data Buffer}: Two data buffers are used to store the real-time data stream awaiting inference and the labeled data prepared for fine-tuning. 
    \item \textbf{Mobile Model Adaptation}: Its frontend performs direct model reuse or fine-tuning based on the mobile model cache. If needed, its backend interacts with the cloud for retrieval from the cloud model database or end-to-end retraining for ultimate adaptation.
\end{itemize}

\subsubsection{\textbf{Cloud Components.}} 
The cloud server stores history models collected from its connected mobile devices and performs computation tasks in the background to assist the mobile device components, 
with the following four modules.
\begin{itemize}[topsep=0pt,leftmargin=0.35cm]
    \item \textbf{Annotator:} A large-scale teacher model is deployed to annotate the mobile-uploaded samples and a multi-modal foundation model is used to recognize the meta-level domain semantics (\ie time, weather, location). 
    \item \textbf{Cloud Model Database:} It stores expert models in historical domains based on predefined semantic dimensions, enabling efficient model retrieval. A taxonomy updating algorithm is designed along with retraining tasks. 
    \item \textbf{Cloud Data Pool:} The labeled samples of historical domains are stored for future domain retraining.
    \item \textbf{Cloud Model Adaptation:} It conducts end-to-end retraining for new domains using data from the cloud data pool and dispatches the trained model to mobile devices.
\end{itemize}



\begin{figure}[t!]
\includegraphics[width=0.75\linewidth]{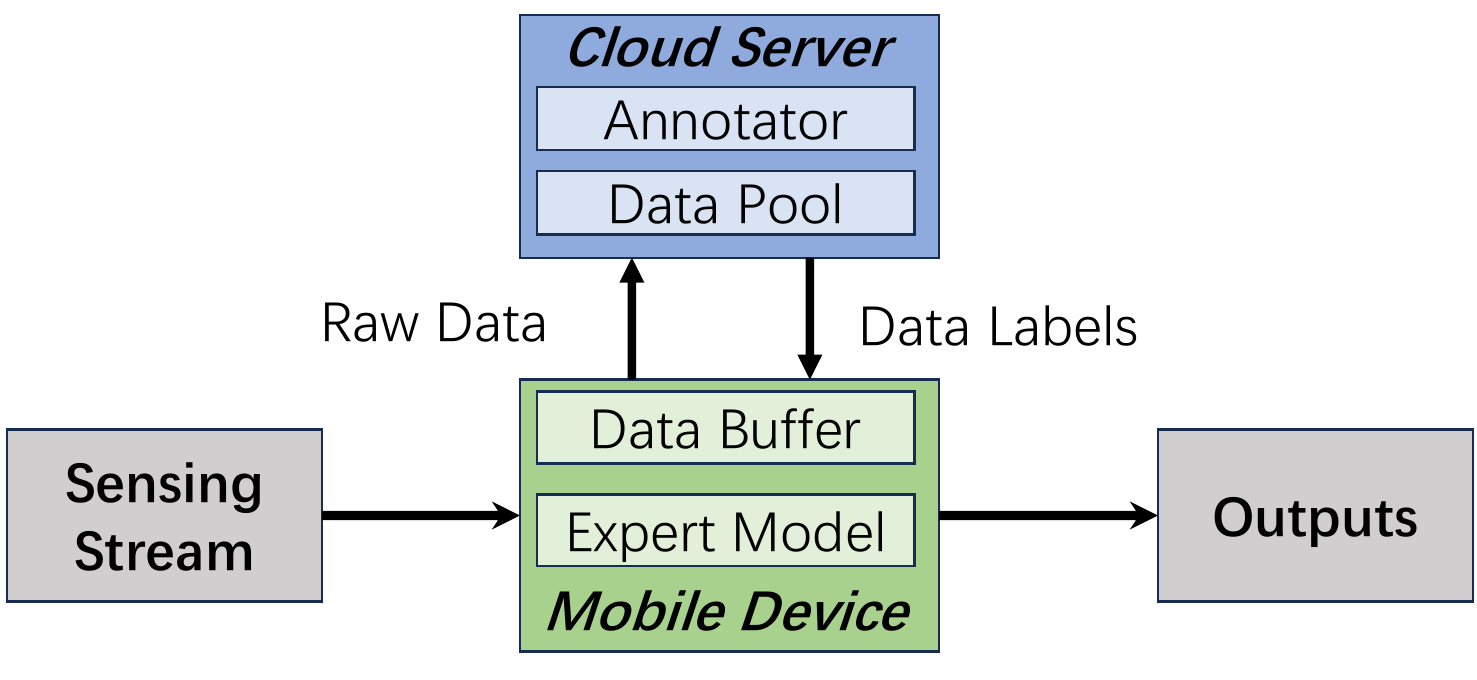}
\caption{Data transmission flow.}
\label{fig:exchange}
\end{figure}

\subsection{Mobile-Cloud Collaboration Workflow} \label{subsec:exchange}
With the introduced mobile and cloud modules, we elaborate on their collaboration procedure during adaptations.
A standard data transmission flow is shown in Figure~\ref{fig:exchange}.
Although the frequency of environment shifts is much lower than the sensor sampling rate (\eg 25 FPS), previous works~\cite{khani2023recl,khani2021real,bhardwaj2022ekya} require continuous data uploading for cloud-based environment shift detection. 
In contrast, \model avoids data transmission during periods without environment shifts by hosting onboard environment shift detection.

This leads to two system states: 
during \textit{``regular inference time''}, only inference tasks and environment shift detection are conducted; 
during \textit{``potential shift time''}, continuous data uploading and annotation are launched to confirm actual environment shifts and to prepare data for adaptations. 
Following existing practices~\cite{khani2023recl,bhardwaj2022ekya}, we set fixed-length \textit{model-update windows} (by default, 30 seconds) to ensure adequate time for fine-tuning data collection on mobile devices and to provide a baseline for waiting times in specific system states. 

As a mobile-oriented framework, most \model actions are triggered and performed on the mobile, while the cloud passively serves the submitted mobile requests.
Within the collaboration, the \textbf{mobile activities} include: 
\begin{enumerate}[leftmargin=0.55cm,topsep=0pt]
    \item It performs regular inference with the deployed expert model on video frames and conducts environment shift detection based on extracted sample features (§\ref{subsec:ood}). If distribution drifts are detected, it enters potential shift time.
    \item In potential shift time, it uploads subsampled frames to the cloud for (\romannumeral1) domain semantic discrimination to confirm the environment shift, and (\romannumeral2) sample annotations for potential adaptations. If the cloud confirms no environment shift happens, the false alarm is resolved and the mobile quits potential shift time. 
    Otherwise, the below adaptation actions are performed.
    \begin{enumerate}[leftmargin=0.55cm,topsep=0pt]
    \item It immediately performs model reuse from the most similar domain(s) to recover accuracy (§\ref{subsec:reuse}) and evaluates the reused model with cloud-annotated samples. If an accuracy threshold is missed, a LoRA fine-tuning task starts accordingly (§\ref{subsec:fine-tune}). 
    \item After step (a), if the new domain has not appeared in the model taxonomy (\ie an unseen domain), its end-to-end training is needed. The mobile continuously uploads samples to the cloud while using the expert model from step (a) for inference. Once enough data is aggregated, the new model is trained and dispatched from the cloud to finish the adaptation. 
    \item After the adaptation sequence, the mobile quits potential shift time and performs model replacement (§\ref{subsec:replace}) in the background for the mobile model cache.
    \end{enumerate}
\end{enumerate}
To assist the mobile adaptations, the \textbf{cloud activities} in each time window include:
\begin{enumerate}[leftmargin=0.55cm,topsep=0pt]
    \item Upon receiving mobile-uploaded samples, it uses the teacher model to annotate task labels and uses the FM to discriminate domain semantics with dedicated prompts. It sends the domain semantics back and stores the annotated samples in its domain buffer.
    \item It serves model retrieval requests from mobile devices by dispatching the expert model to them.
    \item When one window ends, it updates the sample count of each domain and launches end-to-end retraining if enough data is accumulated.
    \item If an end-to-end retraining is finished, the new expert model is stored in its model DB and indexed with domain semantics. The model taxonomy (§\ref{subsec:taxonomy}) is accordingly updated and synchronized with all mobile devices.
\end{enumerate}

\section{Mobile Model Adaptation} \label{sec:fast-adapt}
\begin{figure}[t!]
    \includegraphics[width=\linewidth]{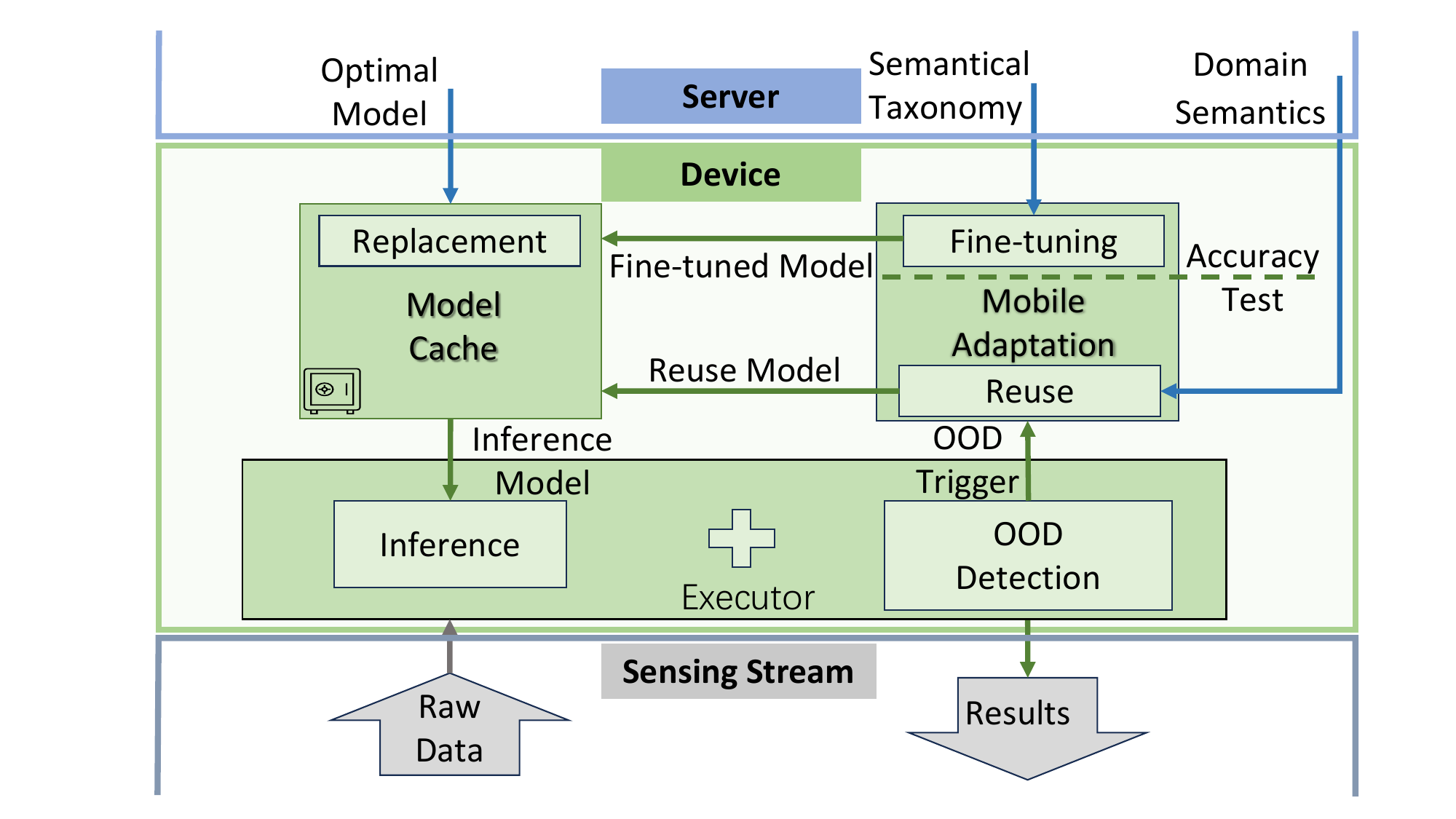}
    \caption{Responsive model adaptation design. 
    }
    \label{fig:cache}
\end{figure}

This section answers four questions related to mobile adaptation components:
(1) how to detect environment shift?
(2) how to load the optimal reuse model? 
(3) how to arrange onboard fine-tuning tasks? 
(4) how to update the cache for future adaptations? 
An overview of the mobile adaptation components is summarized in Figure~\ref{fig:cache}.

\subsection{Environment Shift Detection} \label{subsec:ood}
The end-to-end environment shift detection is achieved through a two-step process.
First, we perform onboard environment shift detection to avoid unnecessary mobile-cloud data transmission during regular inference time. 
Specifically, \model applies a lightweight out-of-distribution (OOD) approach~\cite{lee2018simple} to determine the likelihood of an environment shift without overconfidence bias. 
If an alarm is fired, we switch from regular inference time to potential shift time. 
Second, in potential shift time, the mobile device regularly uploads samples to the cloud to analyze the domain semantics to confirm the actual environment shift. If the environment shift happens, mobile model adaptation strategies are triggered.

We use a multi-dimensional Gaussian distribution to regress the training distribution.
Upon a model training finishes, we choose features from a fixed layer, perform dimensionality reduction with pooling, and record the mean $\hat{\mu}_{c}$ and variance $\hat{\Sigma}$ of the training features.
At runtime detection, the OOD score of a sample is calculated by feeding the extracted sample feature $x_i$ into the equation to get $S(x_{i})$. 
The environment shift indicator of a window $S_{win}(x)$ is based on the aggregated sample OOD scores, as shown below.
\begin{small}
\begin{equation*}
    \hat{\mu}_{c}=\frac{1}{N_{c}}\sum_{i:y_{i}=c}{f(x_{i})}, \quad
    \hat{\Sigma}=\frac{1}{N}\sum_{c}\sum_{i:y_{i}=c}{f(x_{i}-\hat{\mu}_{c})(f(x_{i}-\hat{\mu}_{c}))^{T}},
\end{equation*}
\begin{equation*}
\label{equ:sample_ood}
    S(x_{i})=\max_{c}{[-(f(x_{i})-\hat{\mu}_{c})^{T}\hat{\Sigma}^{-1}(f(x_{i})-\hat{\mu}_{c})]},
\end{equation*}
\begin{equation*}
    S_{win}(x)=\frac{1}{N_{win}}\sum_{0 \leq i < N_{win}}{S(x_{i})},
\end{equation*}
\end{small}
\noindent where $N_{c}$ is the sample count of class $c$, $N$ is the training sample count, and $N_{win}$ is the sample count in a window.


We use the OOD score of point ``$mean+k\times std$'' as the threshold of environment shift detection, with $k$ as a hyper-parameter. 
In §\ref{subsec:micro} we select a $k$ value that is relatively sensitive to distributional drift, to ensure a high recall on the environment shift.
The false positive alarms fired by the mobile environment shift detector could be resolved by the cloud domain discriminator (\ie FM), without initiating meaningless adaptations.
If the cloud FM indicates no environment shift, we exit the potential shift time. 
Such collaborative environment shift detection well balances the environment shift detection efficiency and quality.

\subsection{Mobile Model Reuse} \label{subsec:reuse}
Once an environment shift is confirmed on the cloud, \model performs efficient model reuse as the immediate reaction. Technically, the fitness of a history expert model should be defined by its accuracy in the new domain.
However, annotating samples for the new domain on the cloud and iterating over all history expert models can be time-consuming, we instead use the semantical proximity between domains as an efficient proxy of the model fitness, which is calculated by a distance measure defined on the domain semantical features recognized by the cloud foundation model (as detailed in §\ref{subsec:taxonomy}).
Since the semantical model taxonomy is synchronized on each mobile device, model lookup can be performed locally on each mobile device.

We first identify a ``global optimal model'' on the cloud model database that is expected to achieve the best reuse accuracy. If it is stored in the mobile model cache (\ie cache hit), we directly load it for reuse; otherwise if the ``global optimal model'' is not cached onboard (\ie cache miss), the reuse finishes in two steps. We identify another ``local optimal model'' from the mobile model cache for temporary reuse before the ``global optimal model'' is downloaded from the cloud model database.
In cache hit cases, \model eliminates the time of transmitting data and waiting for the cloud to identify the optimal model. 
Otherwise, it involves cloud-based model dispatch but still saves cloud queueing delays for environment shift detection.
Overall, the mobile model cache and onboard environment shift detector optimize the responsiveness of \model's model reuse.

\subsection{Mobile Model Fine-Tuning} \label{subsec:fine-tune}
Model reuse, although being most responsive, may lead to suboptimal adaptation quality in some cases. 
Thus, we further test the reused expert model with a subset of labeled samples annotated by the cloud. 
If the accuracy misses a threshold (by default 35\% mAP in object detection), at the end of one window, we launch an onboard LoRA fine-tuning task with accumulated labeled samples in the background, without interrupting the inference task. 

Even though mobile fine-tuning takes longer for each iteration than cloud retraining, it achieves completion faster overall as updating single-layer parameters requires magnitudes fewer training samples.
Moreover, fine-tuning avoids cloud queueing delays, while retraining time could increase multiplicatively as more mobile devices are connected to the cloud server. 
In our ablation studies (§\ref{subsec:abla}), fine-tuning is shown to be one critical factor in optimizing the accuracy during the adaptation.
Notably, \model does not upload fine-tuned models to the cloud as it exhibits limited generalizability into other mobile devices.

\begin{figure}[t!]
    \includegraphics[width=0.95\linewidth]{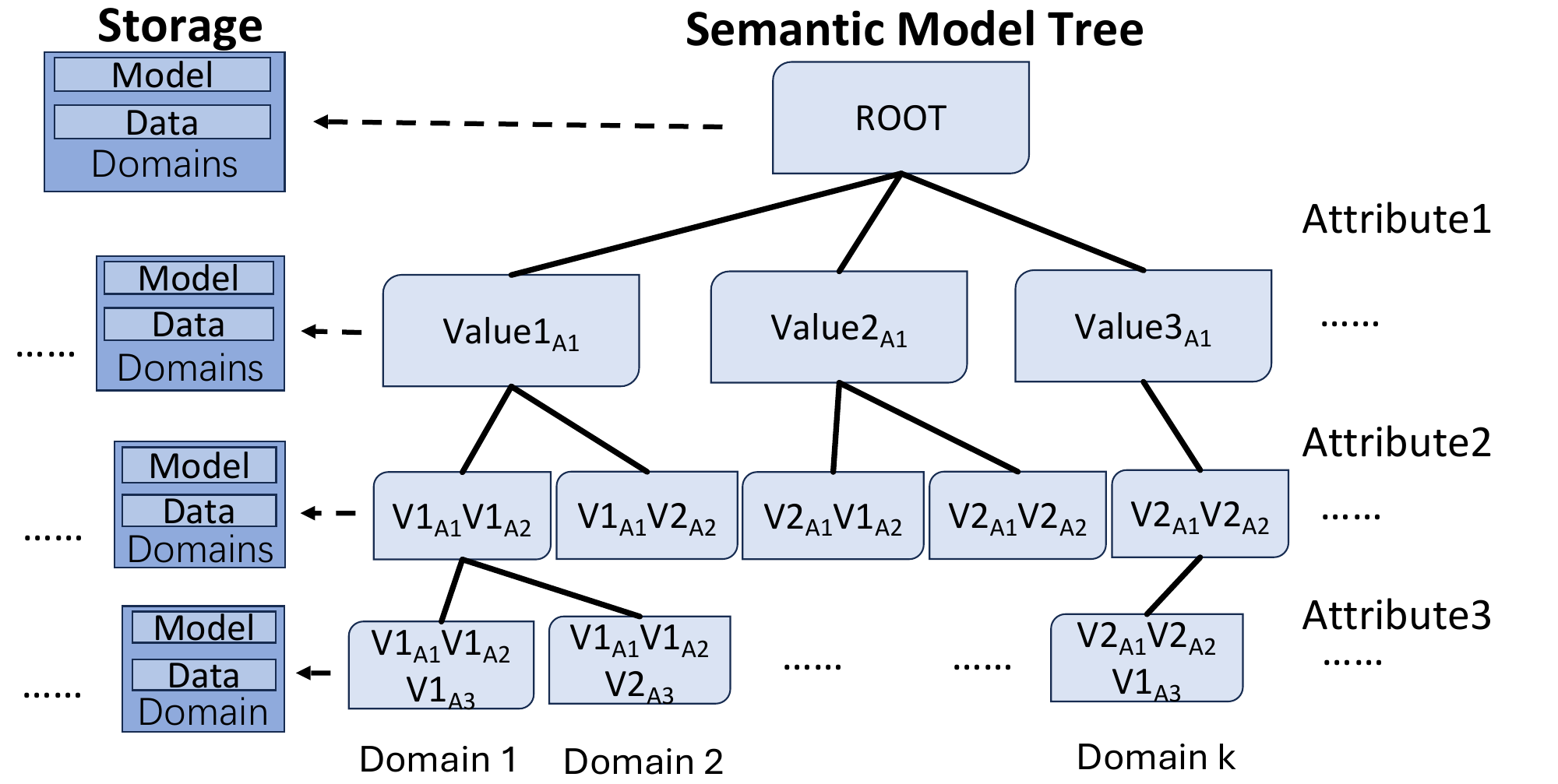}
    \caption{Semantic model taxonomy on the cloud.}
    \label{fig:tree}
\end{figure}

\subsection{Mobile Model Cache Replacement} \label{subsec:replace}
Similar to how the CPU cache prefetches memory pages after a load, we proactively replace expert models in the mobile model cache after reuse to ensure a higher likelihood of cache hits in the next environment shift.
Although future environment shift patterns can be difficult to predict, we choose simple intuitions in cache replacement.

First, the recurrence of environment shift indicates locality in model reuse. For instance, in real-world driving, phenomena such as entering and exiting tunnels or merging onto and off highways are commonly observed. This indicates a pattern of frequent iteration between two domains. Thus, we store the unloaded expert model in the cache for potential use in the next fetch.
Second, within a continuous video stream, the domain can not shift abruptly to semantically distant domains. Instead, similar domains are more likely to happen in the future. Therefore, we prefetch the expert model for the most similar domain to the current domain with the most shared attributes (\eg from ``daytime\_clear\_city-street)'' to ``daytime\_clear\_highway'' or ``daytime\_rainy\_city-street''). 
The cache replacement happens in the background through a best-effort manner without strict deadlines or interrupting the inference task. 

\section{Cloud Model Adaptation} \label{sec:server}
In addition to performing data annotations and domain discrimination upon receiving uploaded samples from mobile, and managing model dispatch in model reuse, the cloud is responsible for maintaining a model taxonomy and serving end-to-end retraining requests.

\subsection{Semantic Model Taxonomy} \label{subsec:taxonomy}
Meta-level domain semantics are used as indices to perform efficient model retrieval and estimate the fitness of cross-domain expert model reuse. 
Our intuition is to organize extensive history expert models into a hierarchical taxonomy rather than a disorganized zoo, such that semantical comparison can be used to replace repetitive model evaluations during selection.
Below we introduce how the domain semantics are organized and utilized.

We request the human operator to specify only the dimensions of semantic attributes in advance (\eg time, weather, and location) while leaving their values open\footnote{The foundation model (FM) possesses the capability of recognizing their values along the given dimensions. Note the semantic dimensions are sorted in the decreasing order of their impact on the model performance, which could be profiled offline with a limited set of scenes.}.
Based on these dimensions, \model generates the domain semantic taxonomy as a tree as visualized in Figure \ref{fig:tree}, where each layer corresponds to an attribute. The child node is more concrete in domain semantics compared to its parent node. The leaf nodes specify values in all attributes and represent most fine-grained domains, while domains at non-leaf nodes cover all their child domains. 
Each node on the semantic model tree has an expert model in the cloud model DB and related labeled data in the cloud data pool.
Besides, we use a meta-level taxonomy table to record the semantical model tree. It is synchronized between the cloud and all mobile devices for model retrieval. 

\begin{figure}[t!]
    \includegraphics[width=0.8\linewidth,]{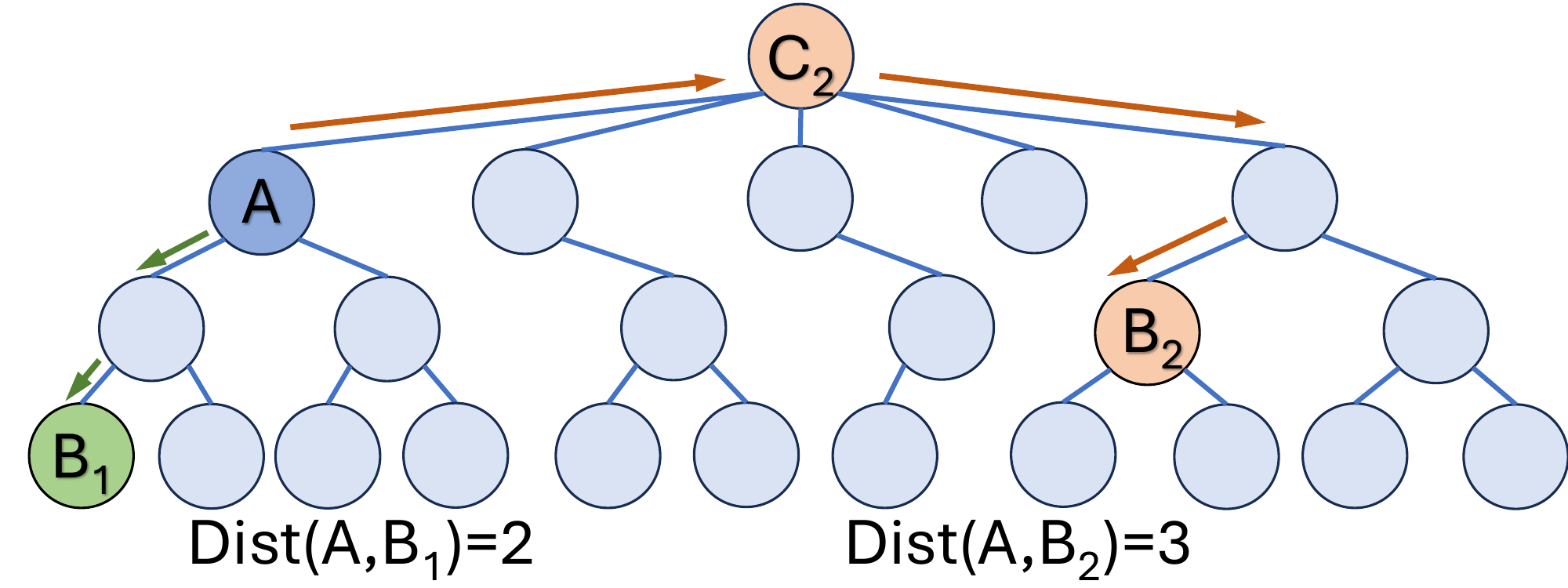}
    \caption{Distance algorithm examples.}
    \label{fig:dist_eg}
\end{figure}

A semantic distance function $Dist$ is defined to measure the similarity of two domains as an indicator of cross-domain expert model reuse accuracy. 
It also assists the expert model prefetch in the mobile model cache management.
The domain semantical distance is defined below.
\begin{equation*}
    Dist(A,B)=\begin{cases}
    |layer(A)-layer(B)|,& A \in S_{B} \vee B \in S_{A}\\
    Dist(A,C)+Dist(B,C),& A \notin S_{B} \wedge B \notin S_{A}\\
    \end{cases}
\end{equation*}
where $A$ and $B$ are the nodes of the tree, $S_{A}$ means the subtree of A, $C$ is the lowest common ancestor node of $A$ and $B$.
We empirically show its effectiveness in proxying the cross-domain performance in §\ref{subsec:micro}. 
Its computation is lightweight (Figure~\ref{fig:dist_eg} as an example) compared to the deep neural network (DNN) based gate networks for expert model selection~\cite{khani2023recl}. 
Meanwhile, with a new domain, the taxonomy table can be easily extended without time-consuming learning. 
Once enough samples are accumulated for a new domain, its model can be end-to-end trained and archived on the cloud. 
We add this new domain to the model taxonomy and synchronize the new taxonomy to all mobile devices.


\begin{figure}[t!]
    \includegraphics[width=0.7\linewidth,]{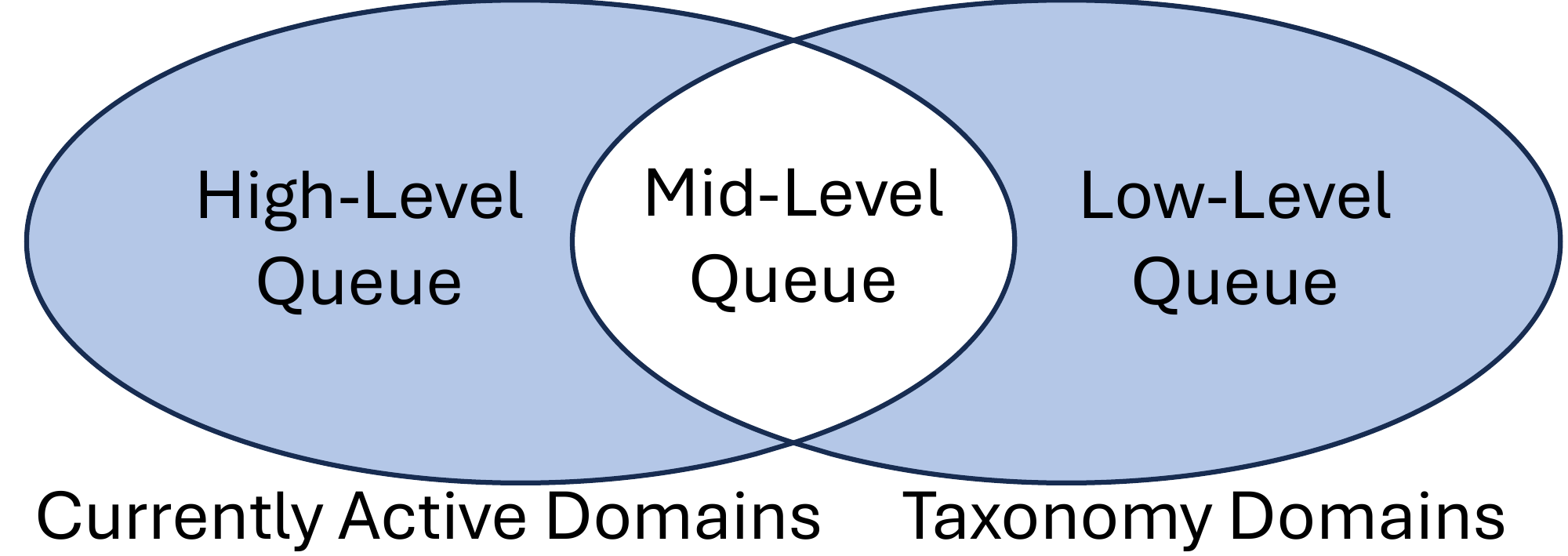}
    \caption{Multi-level queue.}
    \label{fig:mlq}
\end{figure}

\subsection{Cloud Model Retraining} \label{subsec:update} 
In most situations, it is impossible to prepare training data for all domains during the offline training, so we only train models for a subset of domains as the initial point. 
At runtime, as we aggregate enough data for a new domain, we initiate end-to-end training and dynamically expand the model taxonomy.
Meanwhile, an existing domain may meet with plenty of new data shifting from the original data distribution. 
Although fast fine-tuning can partially resolve the in-domain environment shift in some cases, it raises the necessity of end-to-end retraining in other cases with significant shift\footnote{This is regarded as distributional drift caused by unexplainable factors where conventional continuous learning algorithms can fit in to tackle the catastrophic forgetting problem, which is beyond the scope of this paper.}.

We store samples annotated by the teacher model and domain semantics recognized by the FM for new domains and existing domains currently under distribution drift. 
For storage efficiency, we set an upper limit on the sample count per domain (by default, 1000). 
A non-leaf node in the taxonomy consolidates multiple sub-domains, whose (re)training only begins when the node has data from at least two subdomains or a new subdomain has appeared. 
Moreover, the data for the non-leaf domain is sampled in a balanced manner between subdomains for better generalizability. 


\subsection{\textbf{Retraining Task Scheduling.}}
With the number of mobile devices rising, the cloud faces many training tasks to run at a time and each task corresponds to a domain with a set of labeled frames. 
We treat this as a single-processor, non-preemptive scheduling problem~\cite{DBLP:conf/mobicom/000300L00XZWZ24,DBLP:conf/mobicom/LvWL0TYCMGCX24} and apply a multi-level queue scheduling algorithm on the cloud, as shown in Figure~\ref{fig:mlq}. 
Different queues are scheduled with priority scheduling, where retraining tasks in lower-level queues only start when higher-level queues are empty.

Three levels of queues are maintained, and different scheduling policies are applied within each level of task queues.
At the end of each window, two meta-information---the \textit{currently active domain} and \textit{model accuracy in the last window}---are uploaded for each retraining task from mobile devices to determine the corresponding queues they belong to. 
\begin{itemize}[topsep=0pt,leftmargin=0.35cm]
    \item \textbf{High-level queue:} Training tasks of \textit{currently active domains} that have not appeared in the model taxonomy, where an expert model from a similar domain is temporally reused but with suboptimal accuracy. Its tasks are scheduled with a First-In-First-Out (FIFO) policy. 
    \item \textbf{Mid-level queue:} Retraining tasks of currently active domains that are already in the taxonomy. Its tasks are scheduled in the increasing order of \textit{accuracy in the last window}.
    \item \textbf{Low-level queue:} Retraining tasks of domains in the model taxonomy but not currently active, scheduled with FIFO policy. Non-leaf nodes are in the low-level queue because they are only used for temporary reuse.
\end{itemize}

   
\section{Implementation}\label{sec:implementation}
We implemented \model with $6,000$ lines of Python code and implemented all neural networks under PyTorch~\cite{paszke2019pytorch}. The hardware and software setups are specified below.

\begin{figure}[t!]
\includegraphics[width=\linewidth]{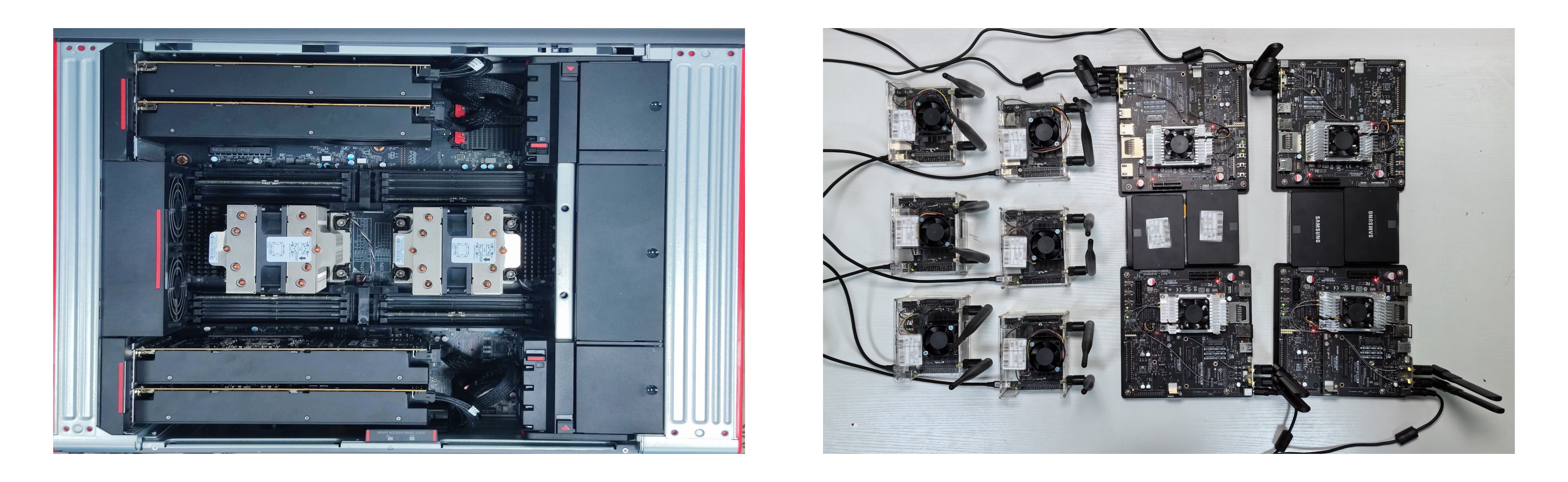}
\caption{Cloud server and mobile devices.}
\label{fig:devices}
\end{figure}

\textbf{Hardware Configurations.} As shown in Figure~\ref{fig:devices}, we use a workstation equipped with an Intel Xeon Gold 5420+ CPU, 128 GB memory, and an NVIDIA RTX 4090 GPU (24 GB RAM) as the cloud server. For mobile devices, we use multiple NVIDIA Jetson TX2, Jetson Nano, and Orange Pi (smartphone-class) boards~\cite{orange_pi_5b}. NVIDIA Jetson devices are configured to MAXN mode to ensure stable performance. Orange Pi boards are set to maximum operating frequency, with RKNN-toolkit~\cite{raul2023rknn} boosting inference speed. Each mobile device runs \model's components alongside a separate input thread to simulate the video stream. 

\textbf{Software Implementation.} We use wondershaper~\cite{wondershaper} to simulate different network conditions (by default, 10Mbps). We use llava-v1.6-vicuna-7b as our FM to discriminate domain semantics. The network data transmission is implemented via TCP Socket APIs. We launch a socket server on the cloud with a fixed port. Each mobile device connects to the port and is allocated an ID stored on the cloud. When transmitting data, the socket API automatically adds its ID to the TCP packages and the cloud can correctly process data from different devices. For retrieval requests, a temporary socket connection is established to transmit model weights without an allocated ID. 
On the mobile side, model inference, model adaptation, and data transmission are executed in separate threads, and the model cache has 3 slots. 
On the cloud server, data annotation, socket connections, and database updates are managed as distinct threads.

\section{Experiments} \label{sec:experiment}
In this section, we report the evaluation results of \model on two video-analytics tasks. 
We first present the methodology (§\ref{subsec:setup}) and end-to-end results (§\ref{subsec:results}), then show micro-performance analysis (§\ref{subsec:micro}), ablation studies (§\ref{subsec:abla}) and mobile overhead (§\ref{subsec:overhead}).


\subsection{Methodology \& Setup} \label{subsec:setup}
\subsubsection{\textbf{Datasets.}} We evaluate \model under 2 different data: domain sequences from BDD dataset~\cite{bdd100k}, and the real-world datasets. \textbf{BDD} dataset consists of 100K $1280\times720$ one-minute-long colored videos collected with dashcams from diverse environments. Ten object classes, along with 108 domains and 3 semantic attributes, are included. We randomly generate 3 domain sequences, and in each domain sequence, we concatenate one video with another video from the same domain when a video ends and change the domain after a fixed number of videos like past frameworks~\cite{suprem13odin,khani2023recl}. By continually changing domains, we simulate different environment shifts and evaluate \model's performance. For image classification, we create the dataset by cropping images from the BDD dataset, using YOLO model annotations to obtain the desired size ($1.4\times$ to the bounding box size) of images with both the domain background and target objects. In these domain sequences, we have the ground truth for when the environment shifts occur, with which we can precisely evaluate the performance of \model's components. 
The \textbf{real-world datasets} consist of videos collected from YouTube and videos collected from our dashcams. All videos are at least an hour long and include various driving environments. Their resolution is $1280\times720$. We use the teacher model for labeling and the label classes are consistent with BDD. We run \model on them to help us evaluate the performance when deployed in natural environments.

\subsubsection{\textbf{Models.}} We leverage YOLOv5-5.0 implementation for object detection, with YOLOv5-s and YOLOv5-x as expert and teacher models respectively. \footnote{The version of Python pre-installed on Nvidia Jetson TX2 is 3.6.9, and higher versions of YOLO are not supported in this environment.} For image classification, we use ResNet implementation in torchvision~\cite{torchvision}, with ResNet18 and ResNet152 as expert and teacher models respectively. For semantic segmentation, we use DeepLabv3+MobileNet and DeepLabv3+ResNet101 as expert and teacher models respectively. We pre-train expert models of 12 random domains and store them in the model database according to configurations to simulate the real-time adaptations.



\subsubsection{\textbf{Metrics.}} To evaluate the system accuracy during continuous adaptation, we compare inference results with labels annotated by the more accurate and more expensive teacher model and ground-truth labels manually annotated from BDD. We use mean Average Precision (mAP) for object detection tasks, classification accuracy (ACC.) for image classification tasks and Mean Intersection over Union (MIoU) for semantic segmentation tasks. In \textbf{end-to-end evaluations}, the following metrics are compared:
\begin{itemize}[leftmargin=0.35cm,topsep=0pt]
    \item \textbf{Recovery accuracy: } To evaluate the adaptation performance of dynamic frameworks during environment shifts instead of several static models with non-adaptation situations, we measure the model accuracy during the \textit{recovery period} from detecting environment shifts to the completion of adaptation. For a fair comparison, we regard the time when the slowest framework, RECL, deploys a new model as the end of recovery, such that all compared frameworks share the same recovery period.
    \item \textbf{Response delay: } It is defined as the duration from when environment shifts are detected to when the first model adaptation action is performed. 
    \item \textbf{Retraining time: }It is defined as the duration from starting an end-to-end retraining on the cloud to deploying the retrained model on the device.
\end{itemize}

\subsubsection{\textbf{Baselines.}} The compared baselines include:
\begin{itemize}[leftmargin=0.35cm,topsep=0pt]
    \item \textbf{No Adaptation}: Pre-trained models are deployed on edge devices without any adaptation.
    
    \item \textbf{ODIN~\cite{suprem13odin}}: It detects and recovers from data drifts based on the similarity of video data and an autoencoder-based model selection. 
    During runtime, the encoder generates an embedding vector for the input data, and ODIN selects the model corresponding to the cluster whose centroid is closest to the embedding vector. In our evaluation, we use the L2 distance as the similarity measure.

    \item \textbf{Ekya~\cite{bhardwaj2022ekya}}: It enables retraining and inference to co-exist on the edge device without model reuse. We utilize Ekya's microprofiler and thief scheduler to manage model retraining jobs for a fair comparison. Despite advanced resource-sharing mechanisms, Ekya experiences out-of-band profiling overhead, which diminishes overall responsiveness.
    
    \item \textbf{RECL~\cite{khani2023recl}}: It enables both retraining and reuse purely on the cloud. We implemented RECL as described in their paper (model selector, safety checker, teacher labeler, and update modules released in RECL). 
    It maintains a gate network for initial model selection, followed by a safety check with a validation dataset.
    Despite its effective model reuse, RECL overlooks the network overhead and focuses much on the update of the gate network.
\end{itemize}

\subsection{End-to-end Evaluation}\label{subsec:results}


\begin{figure*}[t!]
\centering
\begin{minipage}{.45\linewidth}
  \centering
    \includegraphics[width=\linewidth]{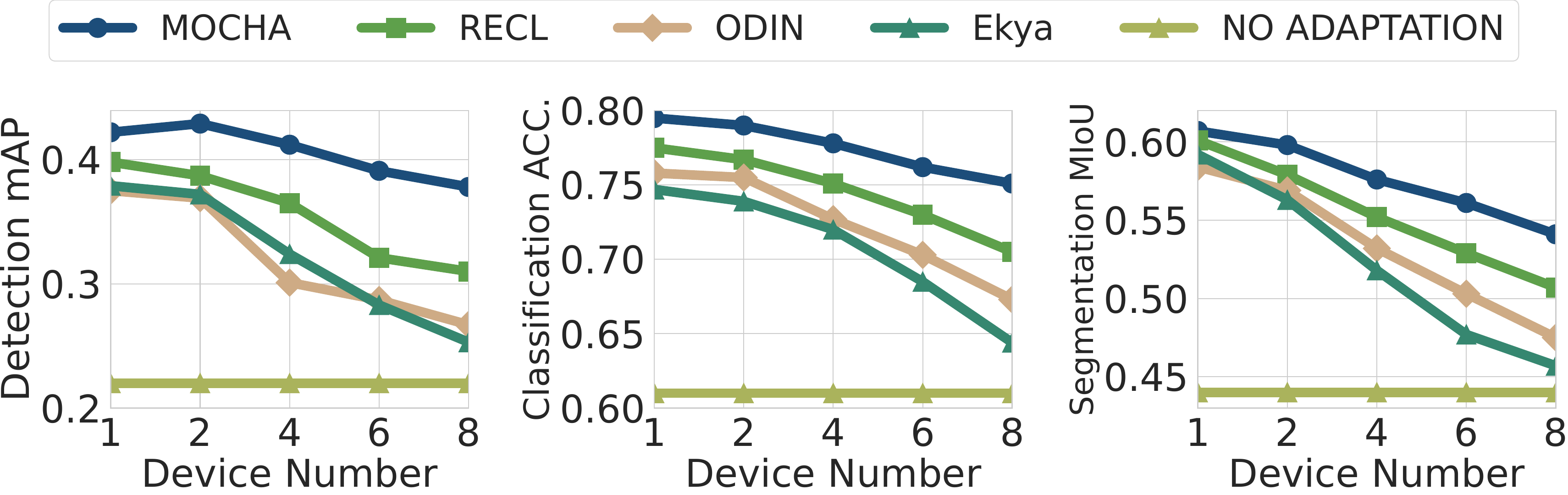}
    \caption{End-to-end evaluation of the recovery accuracy in three tasks.}
    \label{fig:end2end}
\end{minipage}
\begin{minipage}{0.03\linewidth}
\
\end{minipage}
\begin{minipage}{.45\linewidth}
  \centering
    \includegraphics[width=\linewidth]{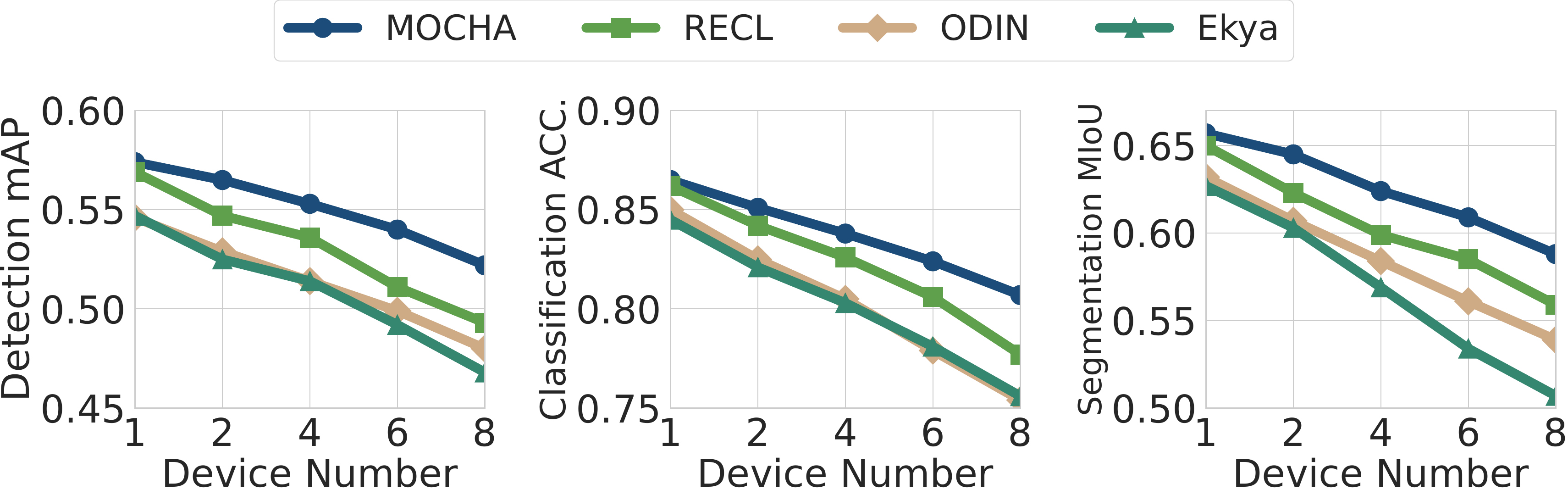}
    \caption{End-to-end evaluation on real-world datasets in three tasks.} 
    \label{fig:real_end2end}
\end{minipage}
\vspace{-0.4cm}
\end{figure*}





\begin{figure*}[t!]
\centering
\begin{minipage}{.45\linewidth}
  \centering
    \includegraphics[width=\linewidth]{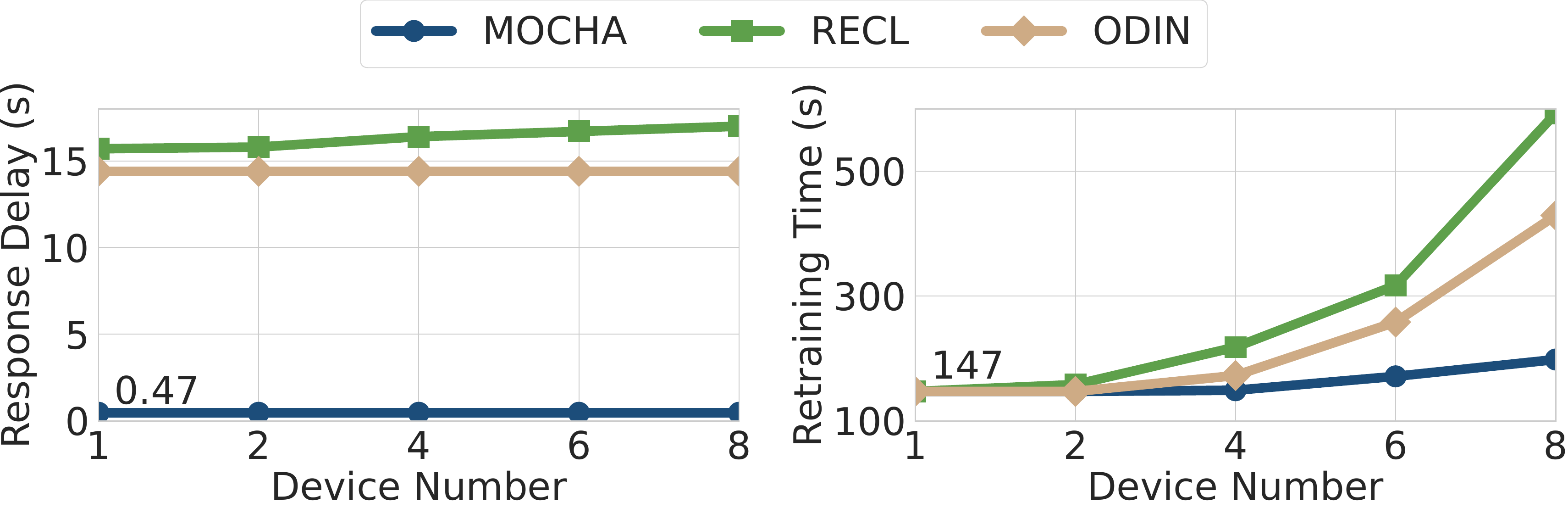}
    \caption{Response delay and retraining time comparison. Object detection task is used.}
    \label{fig:end2end_time}
\end{minipage}
\begin{minipage}{0.03\linewidth}
\
\end{minipage}
\begin{minipage}{.45\linewidth}
  \centering
    \includegraphics[width=0.95\linewidth]{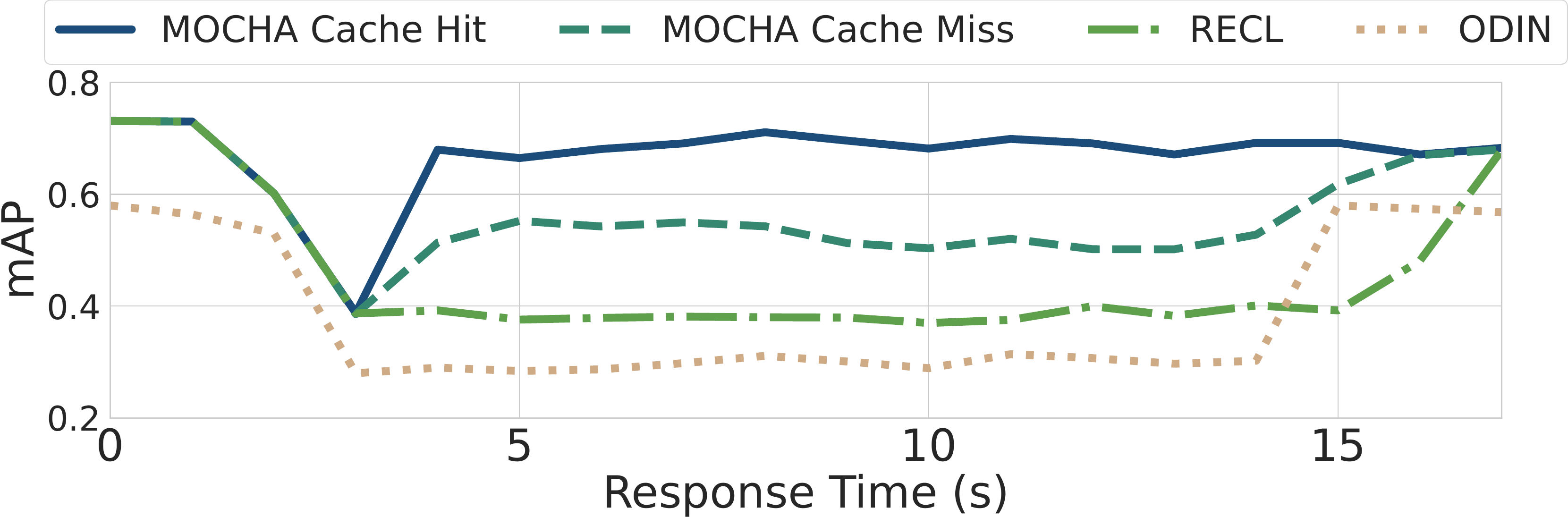}
    \caption{Practical reuse example.}
    \label{fig:reuse_example}
\end{minipage}
\vspace{-0.4cm}
\end{figure*}

\begin{figure*}[t!]
    \includegraphics[width=0.95\linewidth]{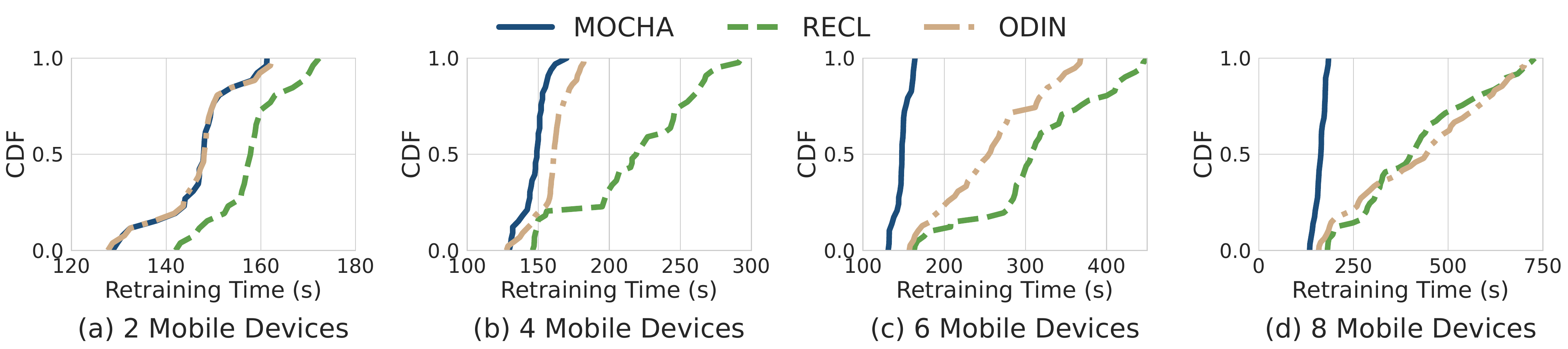}
    \caption{CDF of retraining time across different numbers of devices.}
    \label{fig:od_CDF}
\end{figure*}

\subsubsection{\textbf{Recovery Accuracy.}}
This is the end-to-end quality assessment of all frameworks upon environment shifts.
The recovery accuracy results of \model and the baselines on domain sequences are presented in Figure~\ref{fig:end2end}, with 12 domains included and 3 attributes of semantics used. The recovery accuracy results of four adaptation frameworks on real-world datasets are presented in Figure~\ref{fig:real_end2end}.
\model consistently outperforms all baselines under different numbers of connected mobile devices and different datasets, by up to 6.8\% in the object detection task, up to 4.2\% in the image classification task, and up to 3.8\% in the semantic segmentation task, due to its advanced hierarchical adaptation strategy, allowing faster reuse and efficient fine-tuning. In domain sequences, frameworks will encounter environment shifts with larger variation gaps compared to the real-world datasets we collected. In such more complex scenarios, \model demonstrates a greater advantage over other baselines. 


\subsubsection{\textbf{Response Efficiency.}}
We next compare the response delay between different adaptation algorithms with model reuse in Figure~\ref{fig:end2end_time} and Figure~\ref{fig:reuse_example}.
\model achieves the shortest response delay, $35.5\times$ shorter than the second-best approach. Moreover, \model exhibits better scalability in retraining time when more mobile devices are connected to the cloud.
In Figure~\ref{fig:od_CDF}, \model demonstrates superior robustness in retraining time compared to baselines. It experiences only a modest increase in retraining time as the number of devices grows, resulting in a $1.34\times$ increase. In contrast, RECL shows a $4.03\times$ increase due to the accumulation of ineffective retraining tasks, and ODIN shows a $2.91\times$ increase because of its inefficiency in managing real-world data drift.

\subsubsection{\textbf{Adaptation Analysis.}}
We use example traces to analyze the adaptation collaborations in \model. 
The same domain sequences are evaluated for all frameworks with object detection, and 4 devices are connected to the cloud.

In the \textbf{reuse process} (Figure~\ref{fig:reuse_example}), upon an environment shift, \model's on-device reuse quickly recovers the accuracy through onboard model reuse, while both ODIN and RECL need to wait for the dispatched model from the cloud, creating delays in response. 
``Cache Hit'' and ``Cache Miss'' in \model only differ in whether the expert model for the new domain is cached onboard.
``\model Cache Hit'' finishes with onboard model reuse, providing the best overall accuracy, while ``\model Cache Miss'' partially resolves with onboard cache and goes through the second reuse when the best expert model is dispatched, which still surpasses RECL on the recovery accuracy.
Although ODIN finishes the check and dispatch quickly, it leads to suboptimal recovery accuracy due to the less effective embedding manifold.  

\begin{figure}[t!]
    \centering
    \includegraphics[width=0.95\linewidth]{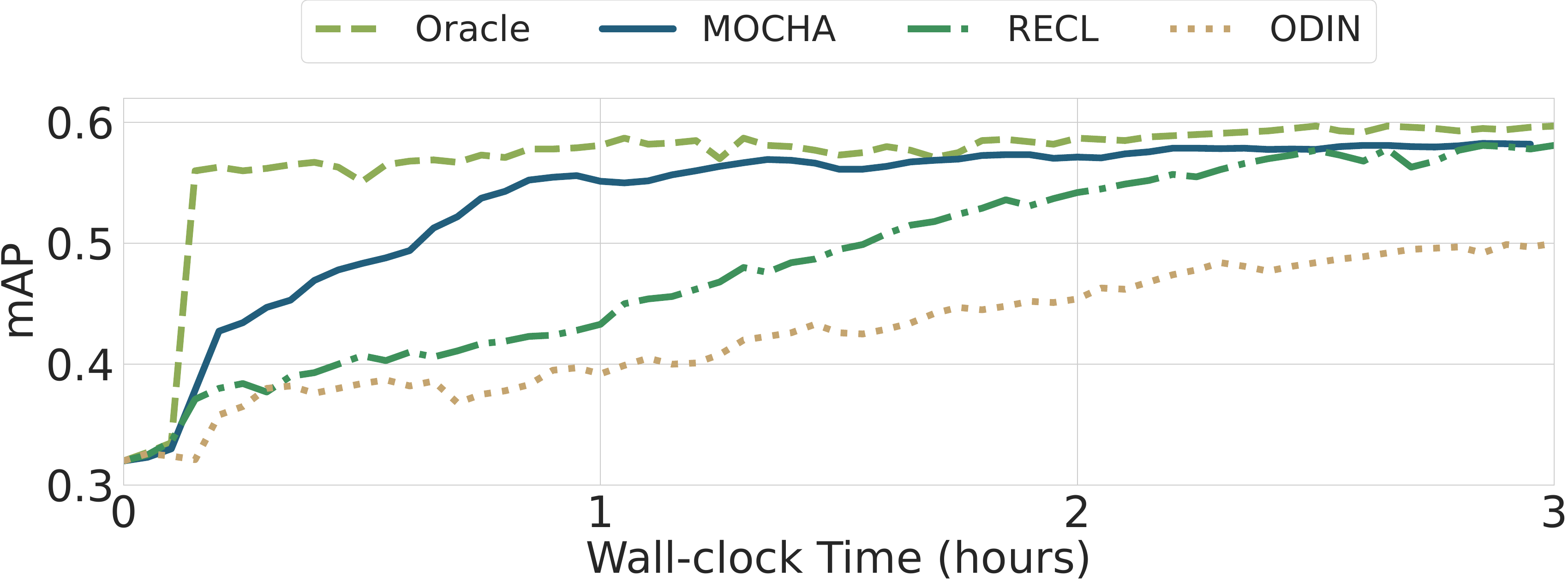}
    \caption{Practical adaptation example over time.}
    \label{fig:wall_clock}
\end{figure}

In Figure~\ref{fig:wall_clock}, we plot the inference accuracy starting from an empty cloud model database, including \model, ODIN, RECL, and the Oracle framework that has sufficient cloud computing resources and bandwidth for multiple retraining tasks. The system ingests 3 hours of video featuring 15 domains. \model exhibits a rapid increase in accuracy due to its superior semantic taxonomy and fine-tuning mechanisms, whereas RECL shows slower improvement. Both \model and RECL can approach Oracle's performance given enough time to expand the model database, while ODIN falls short for a less effective embedding manifold.


\begin{figure}[t!]
    \centering
    \includegraphics[width=0.95\linewidth]{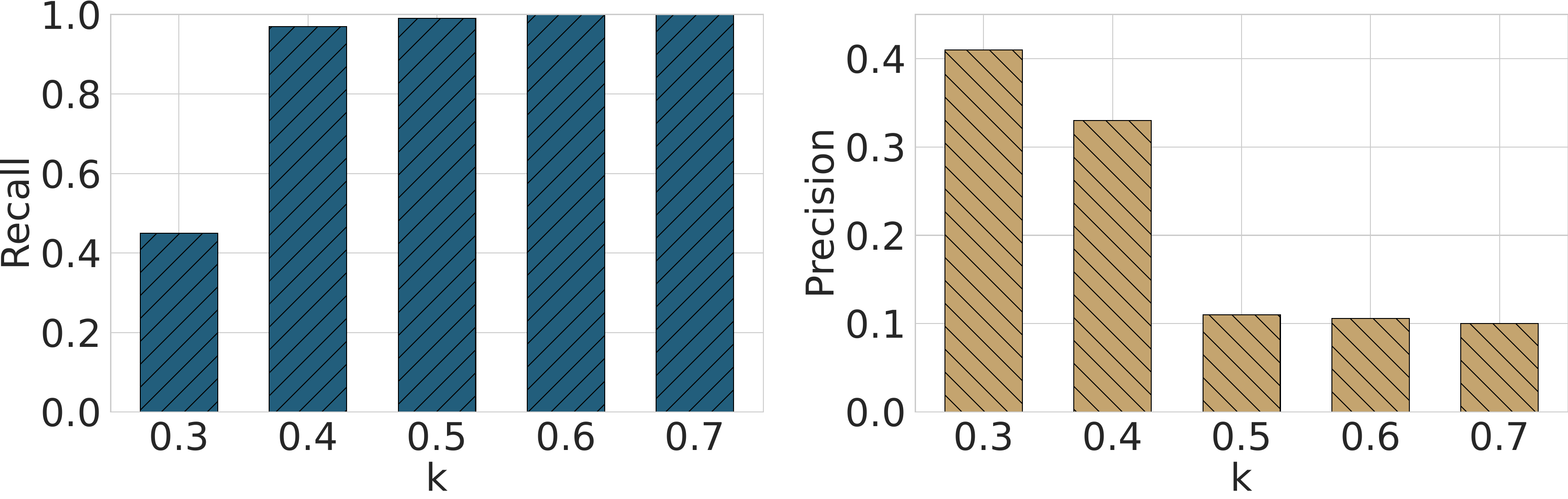}
    \caption{OOD threshold performances by the hyper-parameter $k$.}
    \label{fig:ood_per}
\end{figure}

\begin{figure}[t!]
    \centering
    \includegraphics[width=\linewidth]{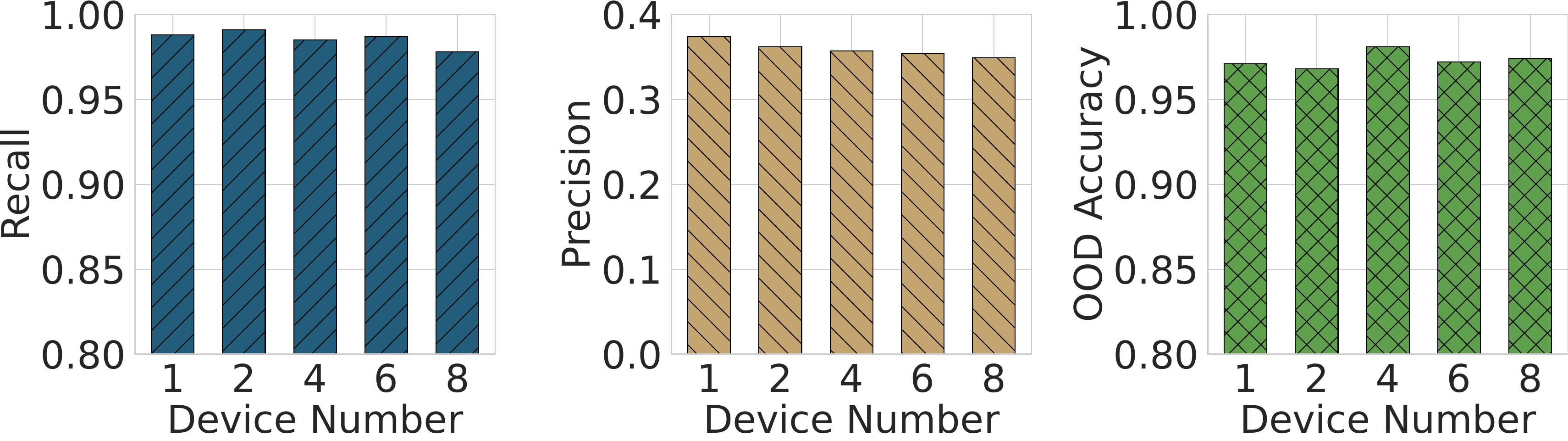}
    \caption{Practical OOD performances on real-world data. OOD accuracy is measured on the §\ref{subsec:results} data with ground truth for environment shifts.}
    \label{fig:real_ood_per}
\end{figure}

\subsection{Micro Experiemnts} \label{subsec:micro}
\subsubsection{\textbf{OOD performance.}}

We compare the precision and recall under different $k$ of the boundary point in environment shift detection (§\ref{subsec:ood}) to choose the best threshold and test the OOD detection accuracy with the data in §\ref{subsec:results}. In Figure~\ref{fig:ood_per}, when $k>=0.4$, the recall exceeds 0.95, which means it is inclusive in detecting potential environment shifts. With $k$ increases, the precision decreases to 0.1, which means OOD triggers false positives without actual environment shifts and \model discriminates these with FM. Thus, ``$k=0.4$'' turns out to be the best configuration. Using the ground truth for when the environment shift occurs, we evaluate the OOD accuracy with the threshold ``$k = 0.4$'' in Figure~\ref{fig:real_ood_per}. The results show that for all environment shifts, an average of 96\% are correctly raised in \model across different numbers of devices, as detailed in the OOD accuracy. The recall and precision values are consistent with those observed in the previous threshold test.



\subsubsection{\textbf{Adaptation ratio vs. Mobile device number}}
\begin{figure}[t!]
    \includegraphics[width=0.95\linewidth]{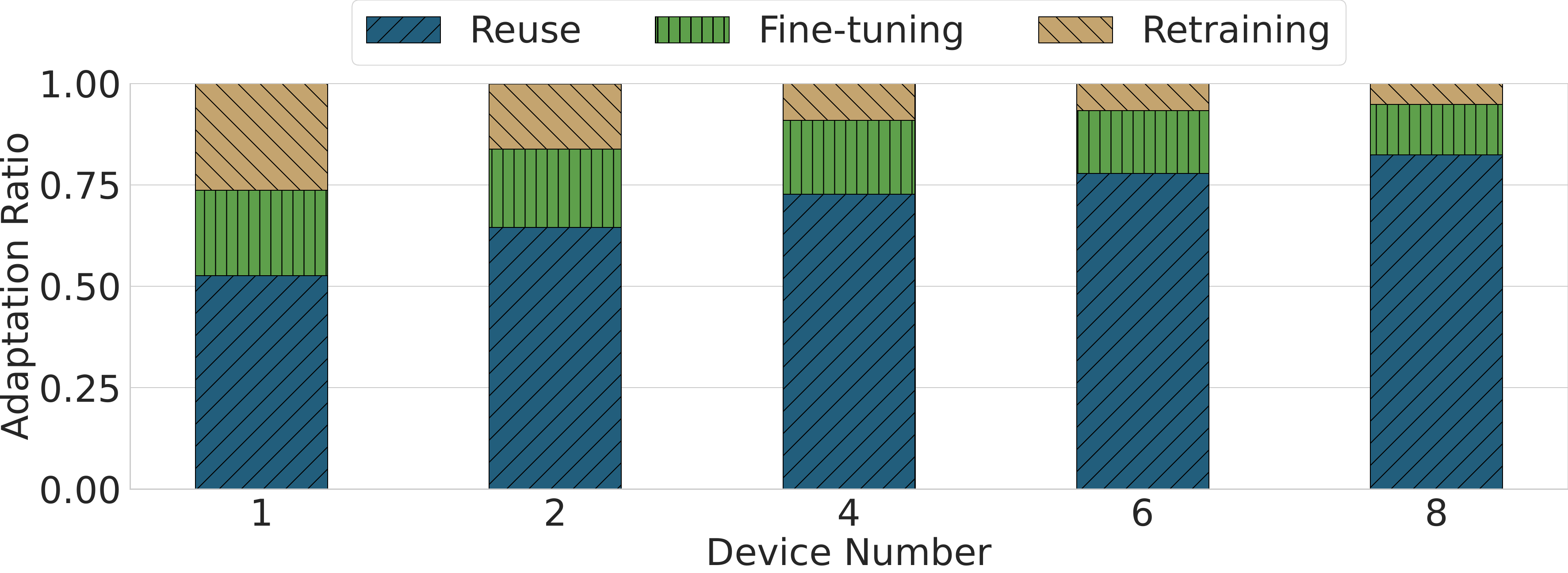}
    \caption{Adaptation ratio changes across different numbers of devices.}
    \label{fig:adapt_ratio}
\end{figure}

As more mobile devices are connected, the adaptation ratios for reuse, fine-tuning, and retraining change, as illustrated in Figure~\ref{fig:adapt_ratio}. With more devices, each device can leverage models in the model DB trained from other devices. This boosts the probability of reuse while reducing the need for fine-tuning and retraining, enhancing the scalability of \model.

\subsection{Ablation Studies} \label{subsec:abla}
\subsubsection{\textbf{Cloud Retraining Scheduler.}}
\begin{figure}[t!]
    \centering
    \includegraphics[width=0.95\linewidth]{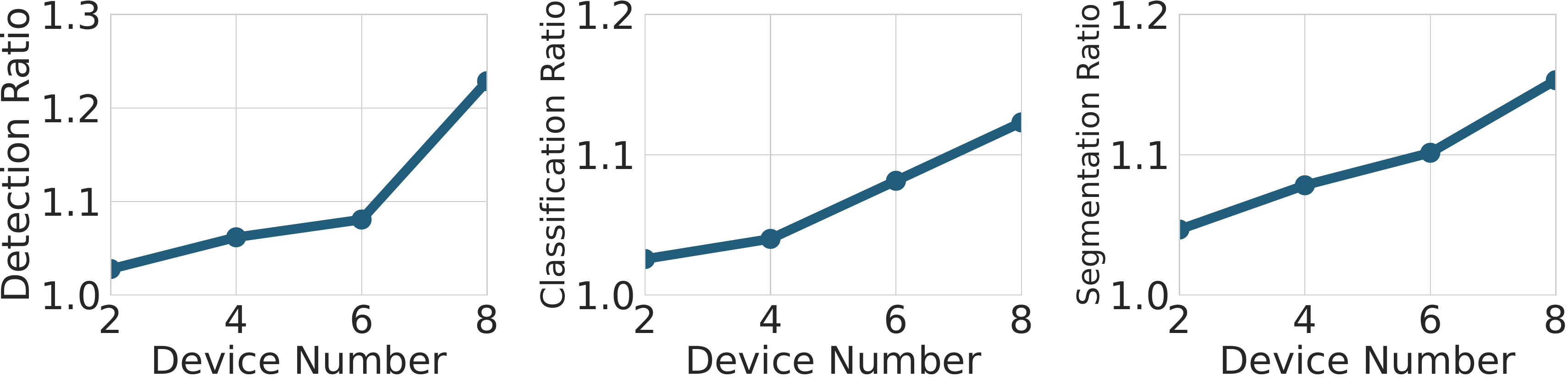}
    \caption{Cloud retraining scheduler comparison. Both ratios represent the comparison of \textit{Multi-Level Queue} (MLQ) over First-In-First-Out (FIFO).}
    \label{fig:sche_per}
\end{figure}

When handling multiple retraining tasks on the cloud, the scheduler performs better using a ``multi-level queue (MLQ)'' algorithm compared to a simple First-In-First-Out (FIFO) approach. We show the accuracy ratios (MLQ/FIFO) in Figure~\ref{fig:sche_per}. With an increasing number of devices, more retraining tasks are parallelized, and the scheduler can optimize task ordering, leading to a better overall performance by up to $1.23\times$ in object detection and $1.12\times$ in image classification.

\subsubsection{\textbf{Mobile model cache.}}
\begin{figure}[t!]
  \centering
    \includegraphics[width=\linewidth]{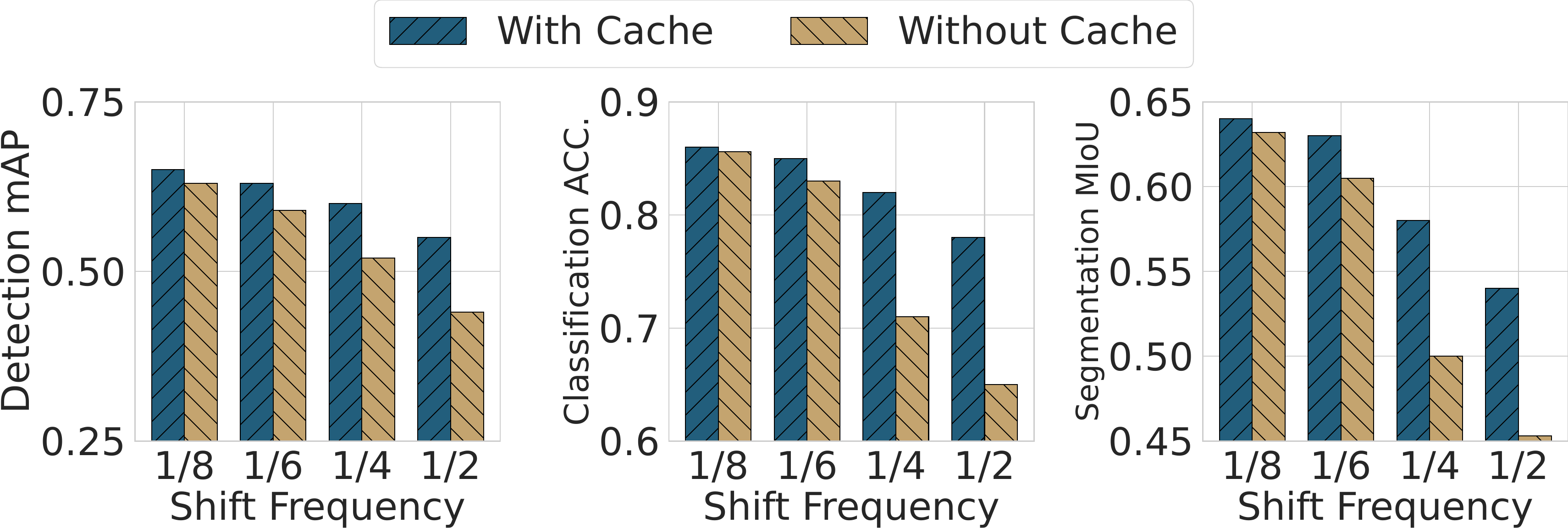}
    \caption{Impact of mobile model cache. Shift frequency represents the reciprocal of the environment shift period, measured in the number of windows.}
    \label{fig:alba_cache}
\end{figure}

In Figure~\ref{fig:alba_cache}, we compare the \model's accuracy with and without the mobile model cache in sequences including frequent environment shifts. The sequences generate environment shifts of existing domains in model DB with fixed numbers of windows and we only launch reuse in \model. Eliminating the model cache, \model experiences an accuracy drop of up to 11\% and 13\% respectively in two tasks. The average reuse time increases to 15.2s and is slightly influenced by the queuing delay on the cloud, which makes \model less competitive in responsiveness. It confirms that the device model cache is the key factor in optimizing the responsiveness of \model.

\subsubsection{\textbf{Model semantic-based selection and fine-tuning.}}
\begin{figure}
    \centering
    \includegraphics[width=0.95\linewidth]{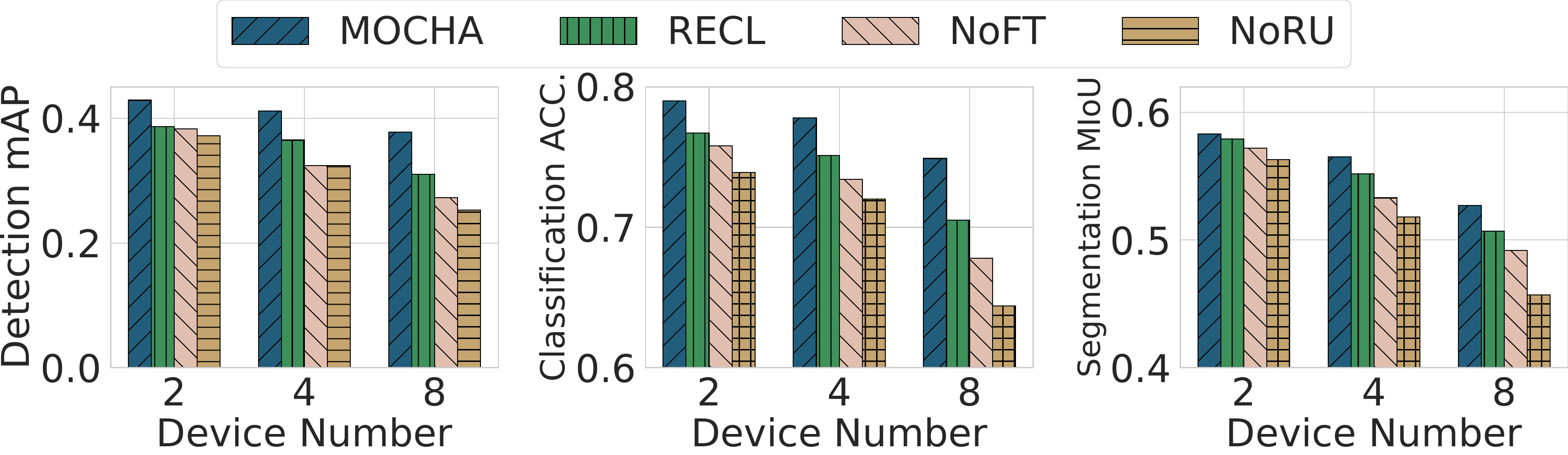}
    \caption{Impact of mobile fine-tuning and mobile selection. Both \model and NoFT use the semantic taxonomy for model selection while RECL uses a gate network. NoFT lacks fine-tuning in \model. NoRU has no model selection or reuse. }
    \label{fig:alba_ft}
\end{figure}

In a model switch process, the traditional approach involves mobile-cloud collaborative model selection and model reuse with a gate network. In contrast, MOCHA implements all model selection, reuse and fine-tuning at the edge. Thus, it is not fair to directly compare \model's lightweight semantic taxonomy with the gate network in model selection. To illustrate the contributions of each step within \model, we compare the performance of four methods: \model, RECL, NoFT(\model without fine-tuning), and NoRU (Ekya without reuse). Results are presented in Figure~\ref{fig:alba_ft}. It is evident that NoRU, utilizing only lightweight model selection, does not have an accuracy advantage over RECL (though it demonstrates significant improvements in responsiveness and scalability). However, the incorporation of reuse allows NoFT to achieve better performance than NoRU. Once fine-tuning is included, \model surpasses RECL in accuracy as well. This is because we are willing to accept a less precise model selection method than the gate network in exchange for responsiveness and lightweight performance while fine-tuning can ensure the accuracy of \model.


\subsection{Mobile Overhead} \label{subsec:overhead}
We measure the inference performance and mobile overhead of \model, RECL, and the ``Inference Only'' baseline in Table~\ref{tab:energy}.
Compared to RECL, during most non-domain-shifting periods without adaptations (without Adap.), \model consumes less power by 7\% and reaches faster speed by 19\% due to the reduced need for frequent data transfers, but extra memory by 10\% because of detection. 
In potential shift time with no fine-tuning, \model consumes the same overhead as RECL, while its power consumption and memory consumption increase by $2.4\times$ and $1.78\times$ respectively, and its inference speed decreases 32\% during the fine-tuning (with Adap.). 
The resource overhead of \model is generally within an acceptable range.

\section{Related Work} \label{sec:related}

\noindent \textbf{Video Analytics System.}  
Video analytics systems aim to achieve high inference accuracy while minimizing energy use and response time through techniques like model distillation~\cite{kang2017noscope,khani2021real,bhardwaj2022ekya,DBLP:conf/mobicom/LiLLCL24}, architecture pruning~\cite{xu2021approxnet,wu2019fbnet,pan2024luoshen,padmanabhan2023gemel}, configuration adaptation~\cite{jain2019scaling,zhang2017live,zhang2018awstream,jiang2018chameleon,meng2023enabling}, frame selection~\cite{li2020reducto,dasari2022swift} and DNN feature reusing~\cite{xu2018deepcache,jiang2018mainstream,hsieh2018focus,guo2021mistify}. The closest to \model is model distillation, where lightweight models are generated to perform on a specific video scene~\cite{kang2017noscope,kang2018blazeit,lee2021benchmarking}. It designs strategies to choose suitable experts based on model retraining and model selection. Retraining techniques train the lightweight models on the latest video frames~\cite{mullapudi2019online,khani2021real,10.1145/3625687.3625800} or on the most relevant images from the training set~\cite{shen2017fast}. Selection techniques maintain and then select a model from a collection of history models~\cite{suprem13odin} or a cascade of models with increasing capacities~\cite{shen2017fast,cao2021thia}. Unlike traditional methods, \model offloads tasks to devices, improving responsiveness and scalability with an on-device model cache and fine-tuning.

\noindent \textbf{Model Selection under data drifts.}
Model selection in a collection of expert models has gained considerable attention, particularly in Mixture-of-Experts (MoE) and autoencoders (AE).
Traditional MoE methods~\cite{riquelme2021scaling,shi2019variational,shazeer2017outrageously,kolter2005using,kolter2007dynamic}, which rely on a gate network, reduce compute costs but necessitate enough memory for loading models and frequent retraining for the gate network. While AE techniques~\cite{suprem13odin,aljundi2017expert,10.1145/3625687.3625785} project input data to a latent space and map new models to a region in the latent space with limited retraining but perform poorly facing environment shifts. In contrast, \model employs an FM and a semantics-based model taxonomy for on-device selection, excelling in few-shot or zero-shot scenarios with minimal updates (§\ref{subsubsec:semantics}).

\noindent \textbf{Continuous learning on the devices.}
In ML literature, CL for mobile devices has been explored in notable works~\cite{hayes2022online,kwon2021exploring} in trade-offs between performance, storage, memory, and compute costs. Recent studies~\cite{bhardwaj2022ekya,ma2023cost} highlight its benefits, motivating offloading adaptations to leverage their natural high response advantages. In \model, on-device reuse and fine-tuning ensure responsiveness and scalability, aligning with our research objectives.

\begin{table}[t!]
\caption{Mobile inference speed and overhead}
\resizebox{\linewidth}{!}{
\begin{tabular}{@{}cc|cccc@{}}
\toprule
\multicolumn{2}{c|}{Methods} & \begin{tabular}[c]{@{}c@{}}Speed\\ (ms/img)\end{tabular} & \begin{tabular}[c]{@{}c@{}}Throughput\\ (FPS)\end{tabular} & \begin{tabular}[c]{@{}c@{}}Memory\\ (GB)\end{tabular} & \begin{tabular}[c]{@{}c@{}}Power\\ (W)\end{tabular} \\ \midrule
\multicolumn{2}{c|}{Inference Only} & 181 & 5.5 & 1.8 & 5.37 \\ \midrule
\multicolumn{2}{c|}{RECL} & 243 & 4.1 & 2.0 & 6.15 \\ \midrule
\multicolumn{1}{c|}{\multirow{2}{*}{MOCHA}} & \multicolumn{1}{c|}{no Adaptation} & 208 & 4.9 & 2.2 & 5.72 \\ \cmidrule(l){2-6} 
\multicolumn{1}{c|}{} & with Adaptation & 354 & 2.8 & 4.8 & 10.96 \\ \bottomrule
\end{tabular}
}
\label{tab:energy}
\end{table}


\section{Conclusion} \label{sec:conclusion}

With cloud offloading being the mainstream solution for continuous mobile video analytics against environment shifts, this paper demonstrated that hierarchical model adaptation through mobile-cloud collaboration offers significant potential in optimizing the responsiveness of adaptation, which proves promising in autonomous driving. 
We presented \model, a mobile-cloud collaborative continuous model adaptation framework, which organically integrates efficient onboard model reuse and fine-tuning with backend model retrieval and retraining on the cloud, achieving improved mobile responsiveness and cloud scalability in our evaluations. 

Still, our current work has limited optimization for replacement strategies of the mobile model cache. And model management on the cloud with numerous devices connected can be further developed to reduce memory overhead and adaptation delays.
We hope our findings can stimulate further research into harnessing the full potential of the synergy between model reuse and model retraining in mobile video analytics systems.

\begin{acks}
This work was sponsored in part by the National Key R\&D Program of China (No. 2022ZD0119100), in part by China NSF grant No. 62472278, 62025204, 62432007, 62441236, 62332014, and 62332013, in part by Alibaba Group through Alibaba Innovation Research Program, and in part by Tencent Rhino Bird Key Research Project. 
This work was partially supported by SJTU Kunpeng \& Ascend Center of Excellence.
The opinions, findings, conclusions, and recommendations expressed in this paper are those of the authors and do not necessarily reflect the views of the funding agencies or the government.
\end{acks}

\newpage 
\bibliographystyle{abbrv}
\bibliography{reference}


\end{document}